%% file: survey.tex
\documentclass[10pt,journal,compsoc]{IEEEtran}

\ifCLASSOPTIONcompsoc
  \usepackage[nocompress]{cite}
\else
  \usepackage{cite}
\fi
\usepackage{amsmath}
\usepackage{amssymb}
\usepackage{amsthm}
\usepackage{graphicx} 
\usepackage{booktabs}
\usepackage{url}
\usepackage{xspace}
\usepackage[table]{xcolor}
\usepackage{pifont}
\usepackage[pagebackref=true,breaklinks=true,bookmarks=false,citecolor=blue,linkcolor=blue,colorlinks,urlcolor=blue]{hyperref}

\input{commands/newcommand}
\input{commands/figure}

\begin{document}

\title{GAN Inversion: A Survey}

\author{Weihao~Xia,
Yulun Zhang,
Yujiu~Yang*, %
Jing-Hao~Xue, %
Bolei~Zhou*,
Ming-Hsuan~Yang*%
\thanks{*Corresponding authors}
\thanks{W.~Xia and Y.~Yang are with Tsinghua Shenzhen International Graduate School, Tsinghua University, China.
Email: xiawh3@outlook.com, yang.yujiu@sz.tsinghua.edu.cn}
\thanks{Y.~Zhang is with the Computer Vision Lab, ETH Z\"{u}rich, Z\"{u}rich 8092, Switzerland.
Email: yulun100@gmail.com}
\thanks{J.-H.~Xue is with the Department of Statistical Science, University College London, UK.
Email: jinghao.xue@ucl.ac.uk}
\thanks{B.~Zhou is with Computer Science Department, University of California, Los Angeles.
Email: bolei@cs.ucla.edu}
\thanks{M.-H.~Yang is with University of California at Merced, Yonsei University, and Google. 
Email: mhyang@ucmerced.edu}
}

\IEEEtitleabstractindextext{
\begin{abstract}
GAN inversion aims to invert a given image back into the latent space of a pretrained GAN model so that the image can be faithfully reconstructed from the inverted code by the generator.
As an emerging technique to bridge the real and fake image domains, GAN inversion plays an essential role in enabling pretrained GAN models, such as StyleGAN and BigGAN, for applications of real image editing. Moreover, GAN inversion interprets GAN's latent space and examines how realistic images can be generated.
In this paper, we provide a survey of GAN inversion with a focus on its representative algorithms and its applications in image restoration and image manipulation. 
We further discuss the trends and challenges for future research.
A curated list of GAN inversion methods, datasets, and other related information can be found at this~\website~site.
\end{abstract}

\begin{IEEEkeywords}
Generative Adversarial Networks, Interpretable Machine Learning, Image Reconstruction, Image Manipulation
\end{IEEEkeywords}
}

\maketitle
\IEEEpeerreviewmaketitle

\section{Introduction}
\label{sec:introduction}
\IEEEPARstart{T}{he} 
Generative Adversarial Network (GAN) is a deep generative model that learns to generate new data through adversarial training~\cite{goodfellow2014generative}. 
It consists of two neural networks: a generator, $G$, and a discriminator, $D$, which are trained jointly through an adversarial process.
The objective of $G$ is to synthesize fake data that resemble real data, while the objective of $D$ is to distinguish between real and fake data. 
Through an adversarial training process, the generator $G$ tries to generate fake data that match the real data distribution to fool the discriminator.
In recent years, GANs have been applied to computer vision tasks ranging from image translation~\cite{isola2017image,zhu2017unpaired,huang2018munit}, image manipulation~\cite{wang2018high,xia2020gaze,li2020manigan}, to image restoration~\cite{zhang2017beyond,tsai2017deep,xu2017text}.

Many GAN models, \eg, PGGAN~\cite{karras2017progressive}, BigGAN~\cite{brock2018large} and StyleGAN~\cite{karras2019style,karras2020analyzing}, have been developed to synthesize images with high quality and diversity from random latent code. 
Recent studies have shown that GANs effectively encode rich semantic information in intermediate features~\cite{bau2019semantic} and latent spaces~\cite{goetschalckx2019ganalyze,jahanian2020steerability, shen2020interpreting} from the supervision of image generation.
These methods can synthesize images with a diverse range of attributes, such as faces with different ages and expressions, and scenes with different lighting conditions. By varying the latent code, we can manipulate certain attributes while retaining the other attributes for the generated image. 
However, such manipulation in the latent space is only applicable to the images generated by the GAN generator rather than any given real images due to the lack of inference capability in GANs. 

\figoverview

GAN inversion aims to invert a given image back into the latent space of a pretrained GAN model. The image can then be faithfully reconstructed from the inverted code by the generator. 
Since GAN inversion plays an essential role in bridging real and fake image domains, significant advances have been made~\cite{zhu2016generative,abdal2019image2stylegan,abdal2020image2stylegan2,bau2019seeing,karras2020analyzing,huh2020transforming,pan2020exploiting,jahanian2020steerability,shen2020interpreting}. 
GAN inversion makes the controllable directions found in latent spaces of the existing trained GANs applicable to editing real images, without requiring any ad-hoc supervision or expensive optimization. 
As shown in Fig.~\ref{fig:overview}, after the real image is inverted into the latent space, we can vary its code along one specific direction to edit the corresponding attribute of the image. 
As a rapidly growing direction that combines GANs and interpretable machine learning techniques, GAN inversion is not only a flexible image editing framework but also helps reveal the inner workings of deep generative models. 

In this paper, we present a comprehensive survey of GAN inversion methods with an emphasis on algorithms and applications. 
To the best of our knowledge, this work is the first survey on the rapidly growing GAN inversion with the following contributions. We provide a comprehensive review of GAN inversion methods and compare their different properties and performances. We further discuss the challenges, open issues, and trends for future research. 

The rest of this survey paper is organized as follows.
We first give a problem formulation of GAN inversion in Section~\ref{sec:definition}.
The obtained latent code for a given image should have two properties: 1) reconstructing the input image faithfully and photorealistically and 2) facilitating downstream tasks. Achieving these two properties is also the goal of GAN inversion. Section~\ref{sec:model_data} introduces many different pretrained GAN models $G(\z)$.
Subsequent sections introduce the efforts taken by different GAN inversion methods to reach the goal.
To evaluate the performance of GAN inversion methods, we consider the two important aspects, how photorealistic (perceptual quality) and faithful (inversion accuracy) the reconstructed image is, in Section~\ref{sec:metrics}.
The first aspect depends on how the formulation is solved. It is usually a nonconvex optimization problem due to the nonconvexity of $G(\z)$, for which finding accurate solutions is difficult. 
The second aspect is primarily decided by which latent space to use. Section~\ref{sec:space} introduces, analyses, and compares the characteristics of different latent spaces. 
In Sections~\ref{sec:techniques}, \ref{sec:characteristics}, and \ref{sec:navigation}, we introduce how existing methods have attempted to provide solutions and discuss some important characteristics of these GAN inversion methods.
Applications and future directions of GAN inversion are introduced in Sections~\ref{sec:applications} and~\ref{sec:outlook}.

\section{Problem Definition and Overview}
\label{sec:definition}

It is well known that GANs~\cite{goodfellow2014generative,karras2017progressive,karras2019style} can generate high-resolution and photorealistic fake images.
However, it remains challenging to apply these unconditional GANs to the editing of real images due to the lack of inference capability.
Given an image, GAN inversion aims to recover the latent code in a latent space of a pretrained unconditional GAN model, and thus enables numerous image editing applications by manipulating the latent code.
In this case, the pretrained unconditional GAN model can be used without modifying the architecture.
Ideally the found latent code of the given image should achieve two goals: 1) reconstructing the input image faithfully and photorealistically and 2) facilitating downstream tasks.

We first define the problem of GAN inversion under a unified mathematical formulation. The generator of an unconditional GAN learns the mapping $G: \mathcal{Z} \to \mathcal{X}$. 
When $\z_1, \z_2 \in \mathcal{Z}$ are close in $\mathcal{Z}$ space, the corresponding images $x_1, x_2 \in \mathcal{X}$ are visually similar. 
GAN inversion maps data $x$ back to latent representation $\z^*$ or, equivalently, finds an image ${x^*}$ that can be entirely synthesized by the well-trained generator $G$ and remain close to the real image $x$.
Formally, denoting the signal to be inverted as $x \in \R^{n}$, the well-trained generator as $G: \R^{n_{0}} \to \R^{n}$, and the latent vector as $\z \in \R^{n_{0}}$, we study the following inversion problem:
\begin{equation}
\z^*=\underset{\z}{\arg \min } \ \ell(G(\z), x),
\label{eqn:def}
\end{equation}
where $\ell(\cdot)$ is a distance metric in the image or feature space, and $G$ is assumed to be a feed-forward neural network. 
Typically, $\ell(\cdot)$ can be based on $\ell_1$, $\ell_2$, perceptual~\cite{johnson2016perceptual} or LPIPS~\cite{zhang2018unreasonable} metrics. Some other constraints on latent codes~\cite{zhu2020indomain} or face identity~\cite{richardson2020encoding} could also be included in practice.
From the obtained $\z^*$, we can obtain the original image; we can vary $\z^*$ to further obtain the manipulated image. %

The second goal as facilitating downstream tasks is primarily decided by which latent space to use (see Section~\ref{sec:space}). 
The first goal depends on how to solve Equation~\eqref{eqn:def} accurately, which is usually a nonconvex optimization problem due to the nonconvexity of $G(\z)$. Thus it is not easily amenable to find accurate solutions. 
Many methods~\cite{zhu2016generative,abdal2019image2stylegan,richardson2020encoding} have been developed to solve Equation~\eqref{eqn:def} with formulation based on learning, optimization, or both.
A {\bf learning-based} inversion method aims to learn an encoder network to map an image into the latent space such that the reconstructed image based on the latent code looks as similar to the original one as possible. 
An {\bf optimization-based} inversion approach directly solves the objective function through back-propagation to find a latent code that minimizes pixel-wise reconstruction loss. 
A {\bf hybrid} approach first uses an encoder to generate initial latent code and then refines it with an optimization algorithm.
Generally, learning-based GAN inversion methods cannot faithfully reconstruct the image content.
For example, learning-based inversion methods have been known to sometimes fail in preserving identities as well as some other details when reconstructing face images~\cite{zhu2020indomain,richardson2020encoding}.
While optimization-based techniques have achieved superior image reconstruction quality, their inevitable drawback is the significantly higher computational cost~\cite{abdal2019image2stylegan,abdal2020image2stylegan2}.
Thus, recent improvements of learning-based GAN inversion methods mainly focus on how to faithfully reconstruct images, \eg,~integrating an additional facial identity loss during training~\cite{richardson2020encoding,wei2021simpleinversion} or proposing an iterative feedback mechanism~\cite{alaluf2021restyle}. 
Recent improvements of optimization-based methods emphasize on how to find the desired latent code more quickly thus propose several initialization strategies~\cite{abdal2019image2stylegan,abdal2020image2stylegan2} and optimizers~\cite{zhu2016generative,huh2020transforming}.
Reconstruction quality and inference time cannot be simultaneously achieved for existing inversion approaches, resulting in a ``quality-time tradeoff''. Although some hybrid approaches are additionally proposed to balance this tradeoff, it remains a challenge to quickly find an accurate latent code. 

Similar to GAN inversion, some tasks also aim to learn the inverse mapping of GAN models.
Some methods~\cite{donahue2016adversarial,dumoulin2016adversarially,lang2021explaining,zhu2019disentangled} use additional encoder networks to learn the inverse mapping of GANs, but their goals are to jointly train the encoder with both the generator and the discriminator, instead of using \textit{a trained GAN model}. 
Some other methods, \eg,~PULSE~\cite{menon2020pulse}, ILO~\cite{daras2021intermediate}, or PICGM~\cite{kelkar2021prior}, also rely on a pretrained generator to solve inverse problems such as inpainting, super-resolution, or denoising. 
They design different optimization mechanisms to search for latent codes that satisfy the given degraded observations. 
Since they aim to search for accurate and reliable estimation (\eg, denoised image) from a degraded observation (\eg, noisy image) instead of \textit{faithful reconstruction of the given image}, we do not categorize them as GAN inversion methods in this survey paper. 
But it would be beneficial to pay attention to those works as they share the same idea of finding desired latent code in the latent space of pretrained GAN models.

\section{Preliminaries}
\label{sec:preliminaries}

\tabfeature

\subsection{GAN Models and Datasets}
\label{sec:model_data}
Deep generative models such as GANs~\cite{goodfellow2014generative} have been used to model natural image distributions and synthesize photorealistic images. Recent advances in GANs, such as DCGAN~\cite{radford2016dcgan}, WGAN~\cite{gulrajani2017improved}, PGGAN~\cite{karras2017progressive}, BigGAN~\cite{brock2018large}, StyleGAN~\cite{karras2019style}, StyleGAN2~\cite{karras2020analyzing}, StyleGAN2-Ada~\cite{Karras2020ada}, and StyleGAN3~\cite{Karras2021free} have developed better architectures, losses, and training schemes. 
These models are trained on diverse datasets, including faces (CelebA-HQ~\cite{karras2017progressive}, FFHQ~\cite{karras2019style,karras2020analyzing}, AnimeFaces~\cite{jin2017towards} and AnimalFace~\cite{liu2019funit}), scenes (LSUN~\cite{yu2015lsun}), and objects (LSUN~\cite{yu2015lsun} and ImageNet~\cite{russakovsky2015imagenet}).
Specifically, BigGAN pretrained on ImageNet, PGGAN on CelebA-HQ, and Style-based GANs on FFHQ or LSUN are widely used in GAN inversion methods.
In contrast to the above-mentioned 2D GANs, the recently developed 3D-aware GANs~\cite{gu2021stylenerf,chan2021pi} bridge the gap between 2D images and 3D physical world. The inversion methods based on these 3D-aware GANs are currently less studied but have great potential for image, video, and 3D applications.

\subsubsection{GAN Models} 
\label{sec:gan models}

\noindent\textbf{DCGAN}~\cite{radford2016dcgan} uses convolutions in the discriminator and fractional-strided convolutions in the generator. \par

\vspace{1mm}
\noindent\textbf{WGAN}~\cite{gulrajani2017improved} minimizes the Wasserstein distance between the generated and real data distributions, which offers more model stability and makes the training process easier.\par

\vspace{1mm}
\noindent\textbf{BigGAN}~\cite{brock2018large} generates high-resolution and high-quality images, with modifications for scaling up, architectural changes and orthogonal regularization to improve the scalability, robustness and stability of large-scale GANs. 
BigGAN can be trained on ImageNet~\cite{russakovsky2015imagenet} at 256$\times$256 and 512$\times$512.\par

\vspace{1mm}
\noindent\textbf{PGGAN}~\cite{karras2017progressive}, also denoted as ProGAN or progressive GAN, uses a growing strategy for the training process. 
The key idea is to start with a low resolution for both the generator and the discriminator and then add new layers that model increasingly fine-grained details as the training progresses. 
This approach improves both the training speed and the stabilization, thereby facilitating image synthesis at higher resolution, \eg, CelebA images at 1024$\times$1024 pixels. \par

\vspace{1mm}
\noindent
\textbf{Style-based GANs}, \eg, StyleGAN~\cite{karras2019style}, implicitly learns hierarchical latent styles for image generation.
This model manipulates the per-channel mean and variance to control the style of an image~\cite{huang2017adain} effectively.
As shown in Fig.~\ref{fig:space}(a), the StyleGAN generator takes style vectors (defined by a mapping network $f$) and stochastic variation (provided by the noise layers) as inputs for image synthesis.
This offers control over the style of generated images at different levels of detail.
The StyleGAN2 model~\cite{karras2020analyzing} further improves the perceptual quality by proposing weight demodulation, path length regularization, generator redesign, and removal of progressive growing. 
The StyleGAN2-Ada~\cite{Karras2020ada} proposes an adaptive discriminator augmentation mechanism to stabilize training with limited data.
StyleGAN3~\cite{Karras2021free} observes an ``texture sticking'' problem (aliasing) in GANs and proposes a new architecture by considering the aliasing effect in the continuous domain and appropriately low-pass filtering the results, which is better suited for video and animation.
For StyleGAN and StyleGAN2, their number of layers $L$ is determined by the output image size $\mathrm{R}$: $L\!=\!2\log_2\mathrm{R}\!-\!2$; it also has a maximum resolution of $1024\times 1024$ with 18 layers.
For StyleGAN3, the number of layers is a free parameter and has no direct relationship to the output resolution.
\par

\figspace

\subsubsection{Datasets}
\label{sec:datasets}

\noindent\textbf{ImageNet}~\cite{russakovsky2015imagenet} is a large-scale hand-annotated dataset for visual object recognition research and contains more than 14 million images with more than 20,000 categories.\par

\vspace{1mm}
\noindent\textbf{CelebA}~\cite{liu2015faceattributes} is a large-scale face attribute dataset consisting of 200K celebrity images with 40 attribute annotations each. CelebA, together with its succeeding CelebA-HQ~\cite{karras2017progressive}, and CelebAMask-HQ~\cite{CelebAMask-HQ}, are widely used in face image generation and manipulation.\par

\vspace{1mm}
\noindent\textbf{Flickr-Faces-HQ} (FFHQ)~\cite{karras2019style} is a high-quality image dataset of human faces crawled from Flickr, which consists of 70,000 high-quality human face images of $1024 \times 1024$ pixels and contains considerable variation in terms of age, ethnicity, and image background.\par

\vspace{1mm}
\noindent\textbf{LSUN}~\cite{yu2015lsun} contains approximately one million labeled images for each of 10 scene categories (\eg, bedroom, church, or tower) and 20 object classes (\eg, bird, cat, or bus).
The church and bedroom scene images and car and bird object images are commonly used in the GAN inversion methods.\par

Some GAN inversion studies also use other datasets
in their experiments, such as \textbf{DeepFashion}~\cite{liu2016deepfashion}, \textbf{AnimeFaces}~\cite{jin2017towards}, and \textbf{StreetScapes}~\cite{naik2014streetscore}.

\subsection{Evaluation Metrics}
\label{sec:metrics}

There are different dimensions to evaluate GAN inversion methods, such as \textit{photorealism}, \textit{faithfulness} of the reconstructed image, and \textit{editability} of the inverted latent code.

\subsubsection{Photorealism}
\label{sec:photorealism}

The IS, FID, and LPIPS metrics are widely used to assess the photorealistic quality of GAN-generated images. 
Other metrics such as Fr$\acute{e}$chet segmentation distance (FSD)~\cite{bau2019seeing} and sliced Wasserstein discrepancy (SWD)~\cite{rabin2011wasserstein} have also been used for image perceptual quality evaluation.
In~\cite{xu2018empirical}, Xu~\etal~present an empirical study on the evaluation metrics of GAN models.

\vspace{1mm}
\noindent
\textbf{Inception score} (IS)~\cite{salimans2016improved} is a widely used metric to measure the quality and diversity of images generated from GAN models. 
It calculates the statistics of a synthesized image using the 
the Inception-v3 Network~\cite{szegedy2016rethinking} pretrained on the ImageNet~\cite{deng2009imagenet}. 
A higher score is better.

\vspace{1mm}
\noindent
\textbf{Fr$\acute{e}$chet inception distance}~\cite{heusel2017gans} (FID) is defined by the Fr$\acute{e}$chet distance between feature vectors from the real and generated images based on the Inception-v3~\cite{szegedy2016rethinking} pool3 layer.
Lower FID indicates better perceptual quality.

\vspace{1mm}
\noindent
\textbf{Learned perceptual image patch similarity} (LPIPS)~\cite{zhang2018unreasonable} measures image perceptual quality using a VGG model~\cite{simonyan2014very} pretrained on the ImageNet. A lower value means higher similarity between image patches.

\subsubsection{Faithfulness}
\label{sec:faithfulness}

Faithfulness measures the similarity between the real image and the generated one. It can be approximated by the image similarity.
The most widely used metrics are PSNR and SSIM.
Some methods use the pixel-wise reconstruction distance, \eg,~mean absolute error (MAE), mean squared error (MSE), or root mean squared error (RMSE).

\vspace{1mm}
\noindent\textbf{Peak signal-to-noise ratio} (PSNR) is one of the most widely used criteria to measure the quality of reconstruction.
The PSNR between the ground truth image and the reconstruction is defined by the maximum possible pixel value of the image and the mean squared error between images.

\vspace{1mm}
\noindent\textbf{Structural similarity} (SSIM)~\cite{TIP2004ImageWang} measures the structural similarity between images based on independent comparisons in terms of luminance, contrast, and structures.
The details of these terms can be found in~\cite{TIP2004ImageWang}.

\subsubsection{Editability}
\label{sec:editability}

Editability measures the editable flexibility of the inverted latent code with respect to certain attributes of the output image from the generator. 
Directly evaluating editability of the latent code is difficult. Existing methods use either cosine or Euclidean distance~\cite{nitzan2020harness} or classification accuracy~\cite{voynov2020latent} to evaluate certain attributes between input $x$ and output $x^{\prime}$ (\ie, modifying the target attribute while keeping others unchanged).
Existing methods focus on evaluation of editability on face data and facial attributes.
For example, Nitzan~\etal~\cite{nitzan2020harness} use the cosine similarity to compare the accuracy of facial expression preservation, which is calculated by the Euclidean distance between 2D landmarks of $x$ and $x^{\prime}$. 
In contrast, the pose preservation is calculated as the Euclidean distance between Euler angles of $x$ and $x^{\prime}$.
Abdal~\etal~\cite{abdal2020styleflow} develop the edit consistency score (regressed by an attribute classifier) to measure the consistency across edited face images based on the assumption that different permutations of edits should have the same attribute score when classified with an attribute classifier.
These methods measure preservation of face identity to evaluate the quality of the edited images.
We note the above-discussed methods may not be applicable to all image domains other than faces.

\subsubsection{Subjective Metric}
\label{sec:subjective}

Aside from the above-mentioned metrics, some studies~\cite{abdal2020styleflow,zhu2020improved} include human raters or user studies for performance evaluation. 
For example, for subjective image quality assessment, human raters are asked to assign perceptual quality scores to images, \eg, from $1$ (bad) to $5$ (good). 
The final score, usually called the mean opinion score (MOS) or difference mean opinion score (DMOS), is calculated as the arithmetic mean over all ratings.
A typical user study asks participants to choose one that best meets the question from a given triple of images (source, results of a baseline and the proposed method).
The question can be ``choose one from the given two edited images that better preserves the identity of the person in the source image'' or ``which edited image is more realistic?''
The final percentage of responses indicates the preference rate of the proposed method against a baseline.
Drawbacks with these metrics include the nonlinear scale of human judgement, potential bias and variance, and high human cost.

\section{GAN Inversion Methods}
\label{sec:model}

This section introduces different latent spaces of GAN models, representative GAN inversion methods, and their properties.
As the StyleGAN models achieve state-of-the-art image synthesis, numerous GAN inversion methods have been developed using various latent spaces~\cite{karras2019style,karras2020analyzing,Karras2021free} based on the StyleGANs.
In addition to the $\mathcal{Z}$ space for generic GANs, several latent spaces are designed specifically for StyleGANs, including ~$\mathcal{W}$, $\mathcal{W}^{+}$, $\mathcal{S}$, and $\mathcal{P}$ spaces.

\subsection{Which Space to Embed - From $\mathcal{Z}$ Space to $\mathcal{P}$ Space}
\label{sec:space}

Regardless of the GAN inversion methods, one important design choice is to which latent space to embed the image. 
A good latent space should be disentangled and easy to embed.
The latent code in such a latent space has the following two properties: it reconstructs the input image faithfully and photorealistically, and it facilitates downstream image editing tasks. 
This section introduces the efforts of latent space analysis and regularization on the latent spaces from the original $\mathcal{Z}$ space to the most recent $\mathcal{P}$ space. 
The $\mathcal{Z}$ space is applicable to  all GANs and some latent spaces are designed specifically for StyleGANs~\cite{abdal2019image2stylegan,wu2020stylespace,zhu2020improved,bai2022high}.
The choice of latent space depends on the pretrained models and tasks. For instance, image editing with StyleGANs is mostly performed in the $\mathcal{W}^{+}$ space.

\vspace{1mm}
\noindent\textbf{$\mathcal{Z}$ Space.} The generative model in the GAN architecture learns to map the values sampled from a simple distribution, \eg, normal or uniform distribution, to the generated images. 
These values, sampled \emph{directly} from the distribution, are often called latent codes or latent representations (denoted by $\z \in \mathcal{Z}$), as shown in Fig.~\ref{fig:space}.
The structure they form is typically called latent $\mathcal{Z}$ space. 
The $\mathcal{Z}$ space is applicable to all the unconditional GAN models such as DCGAN~\cite{radford2016dcgan}, PGGAN~\cite{karras2017progressive}, BigGAN~\cite{brock2018large}, and StyleGANs~\cite{karras2019style,Karras2020ada,karras2020analyzing}.
However, the constraint of the $\mathcal{Z}$ space subject to a normal distribution limits its representation capacity and disentanglement for the semantic attributes. 

\vspace{1mm}
\noindent\textbf{$\mathcal{W}$ and $\mathcal{W}^{+}$ Space.}
Recent GAN inversion methods mostly adopt the latent spaces used in StyleGANs. These latent spaces have higher degrees of freedom and thus are significantly more expressive than the $\mathcal{Z}$ space. Fig.~\ref{fig:space} illustrates the latent spaces from which the inversion methods are constructed. 
Various latent spaces are derived from the original $\mathcal{Z}$ space.
StyleGAN~\cite{karras2019style} converts native $\z$ to the mapped style vectors $\w$ by a nonlinear mapping network $f$ implemented with an $8$-layer multilayer perceptron (MLP). 
This intermediate latent space is named as $\mathcal{W}$ space.
Due to the mapping network and affine transformations, the $\mathcal{W}$ space of StyleGAN contains more disentangled features than does the $\mathcal{Z}$ space. 
Some studies~\cite{shen2020interpreting,abdal2019image2stylegan} analyze the separability and semantics of both $\mathcal{W}$ and $\mathcal{Z}$ spaces.
The expressiveness of $\mathcal{W}$ space is, however, still limited, restricting the range of images that can be faithfully reconstructed. 
Therefore, some works~\cite{abdal2019image2stylegan,abdal2020image2stylegan2} make use of another layer-wise latent space, $\mathcal{W}^{+}$, where a different intermediate latent vector, $\w$, is fed into each of the generator's layers via AdaIN~\cite{huang2017adain}. 
However, inverting images into the $\mathcal{W}^{+}$ space alleviates distortion at the expense of compromised editability.
Recent methods~\cite{tov2021designing,alaluf2021hyperstyle} aim to balance the reconstruction-editability tradeoff by predicting latent codes in $\mathcal{W}^{+}$ that reside close to $\mathcal{W}$.
For a StyleGAN with 18 layers, $\w \in \mathcal{W}$ has 512 dimensions, and $\w \in \mathcal{W}^{+}$ has 18$\times$512 dimensions.

\vspace{1mm}
\noindent\textbf{$\mathcal{S}$ Space.} 
The style space $\mathcal{S}$~\cite{wu2020stylespace} is  spanned by channel-wise style parameters \textit{s}, where \textit{s} is transformed from $\w\in\mathcal{W}$ by using a different learned affine transformation for each layer of the generator.
In a 1024$\times$1024 StyleGAN2 with 18 layers, $\mathcal{W}$, $\mathcal{W}^{+}$, and $\mathcal{S}$ have 512, 9216, and 9088 dimensions, respectively.
This $\mathcal{S}$ space is proposed to achieve better spatial disentanglement in the spatial dimension beyond the semantic level. 
The spatial entanglement is primarily caused by the intrinsic complexity of style-based generators~\cite{karras2019style} and the spatial invariance of AdaIN normalization~\cite{huang2017adain}. 
Xu~\etal~\cite{xu2020hierarchical} replace original style codes with disentangled multilevel visual features learned by an encoder. They refer to the space spanned by these style parameters as $\mathcal{Y}$ space, but it actually can be seen as a type of $\mathcal{S}$ space.
By directly intervening the style code $s \in \mathcal{S}$, methods~\cite{wu2020stylespace,patashnik2021styleclip} based on $\mathcal{S}$ space achieve fine-grained controls on local translations.

\vspace{1mm}
\noindent\textbf{$\mathcal{P}$ Space.} A recent method, PULSE~\cite{menon2020pulse}, has observed a ``soap bubble'' effect when searching a generative model's latent space to find the desired points. As indicated by the name, the ``soap bubble'' effect is that much of the density of a high-dimensional Gaussian lies close to the surface of a hypersphere. The above authors propose embedding images onto the surface of a hypersphere in $\mathcal{Z}$ space. 
Based on the observation, Zhu~\etal~\cite{zhu2020improved} propose a $\mathcal{P}$ space.
Since the last leaky ReLU uses a slope of 0.2, the transformation from $\mathcal{W}$ space to $\mathcal{P}$ space is $\x = \operatorname{LeakyReLU_{5.0}}(\w)$, where $\w$ and $\x$ are latent codes in $\mathcal{W}$ and $\mathcal{P}$ space, respectively.
They make the simplest assumption that the joint distribution of latent codes is approximately a multivariate Gaussian distribution and further propose $\mathcal{P_N}$ space to eliminate the dependency and remove redundancy. 
The transformation from $\mathcal{P}$ space to $\mathcal{P_N}$ space is obtained by PCA whitening: $\hat{\mathbf{v}}=\mathbf{\Lambda}^{-1} \cdot \mathbf{C}^{T}(\mathbf{x}-\boldsymbol{\mu})$, where $\mathbf{\Lambda}^{-1}$ is a scaling matrix, $\mathbf{C}$ is an orthogonal matrix, and $\mu$ is a mean vector. The parameters $\mathbf{C}$, $\boldsymbol{\Lambda}$, and $\boldsymbol{\mu}$ are obtained from $\operatorname{PCA}(\mathbf{X})$, in which $\mathbf{X} \in \mathbb{R}^{10^{6} \times 512}$ consists of 1 million latent samples in $\mathcal{P}$ space.
Such transformation normalizes the distribution to be of zero mean and unit variance, leading to the $\mathcal{P}$ space being isotropic in all directions.
The $P_{N}^{+}$ space is extended from $P_{N}$ space:
$\mathbf{v}=\left\{\mathbf{\Lambda}^{-1} \mathbf{C}^{T}\left(\mathbf{x}_{i}-\boldsymbol{\mu}\right)\right\}_{i=1}^{18}$.
Each of the latent codes is used to demodulate the corresponding StyleGAN feature maps at different layers. 

\subsection{GAN Inversion Methods}
\label{sec:techniques}
Fig.~\ref{fig:inversion_types} shows three main techniques of GAN inversion, \ie, projecting images into the latent space based on learning, optimization, or hybrid formulations.
The inverted codes have other properties, \ie, having supported resolution, being semantic-aware, being layerwise, and having out-of-distribution generalizability.
Table~\ref{tab:taxonomy} lists some important properties of the existing GAN inversion methods.

\subsubsection{Learning-based GAN Inversion}
\label{sec:learning-based}
Learning-based GAN inversion~\cite{perarnau2016invertible,zhu2016generative,bau2019inverting} typically involves training an encoding neural network $E(x; \theta_E)$ to map an image, $x$, into the latent code $\z$ by

\begin{equation}
\theta_E^* = \underset{\theta_E}{\arg\min} \sum_{n} \mathcal{L} (G(E(x_n; \theta_E)), \,x_n),
\label{eqn:rec_train}
\end{equation}
where $x_n$ denotes the $n$-th image in the dataset.
The objective in~\eqref{eqn:rec_train} is reminiscent of an autoencoder pipeline, with an encoder $E$ and a decoder $G$. The decoder $G$ is fixed throughout the training.
Aside from accurate reconstruction, a \textit{good} encoder for GAN inversion should have the following feats: 1) lightweight; 2) data-efficiency; 3) supporting high-resolution images (see Section~\ref{sec:resolution}); and 4) generalizability to arbitrary images (see Section~\ref{sec:ood}).

\figtype

One earlier learning-based GAN inversion method is proposed by Perarnau~\etal~\cite{perarnau2016invertible}. 
Given a conditional GAN (cGAN) model,  a real image $x$ is encoded by a latent code $\z$ and an attribute vector $y$, a modified image $x^{\prime}$ is synthesized by changing $y$. 
This approach consists of training an encoder $E$ with a trained conditional GAN (cGAN). 
Different from Zhu~\etal~\cite{zhu2016generative}, this encoder $E$ is composed of two modules: $E_{z}$, which encodes an image to $\z$, and $E_{y}$, which encodes an image to $y$.
To train $E_{z}$, this method uses the generator to create a dataset of generated images $x^{\prime}$ and  latent vectors $\z$, minimizes a squared reconstruction loss $\mathcal{L}_{ez}$ between $\z$ and $E_{z}(G(\z, y^{\prime}))$ and improves $E_{y}$ by directly training with $\|y-E_{y}(x)\|_{2}^{2}$. 
$E_{y}$ is initially trained by using generated images $x^{\prime}$ and their conditional information $y^{\prime}$.

Due to the prevalence of StyleGANs~\cite{karras2019style,karras2020analyzing,Karras2020ada,Karras2021free}, most recent learning-based methods design an encoder for StyleGANs. 
Richardson~\etal~\cite{richardson2020encoding} propose the \textsc{map2style} modules to learn styles from the corresponding feature map, where 18 single-layer latent codes are predicted separately.
Instead of using 18 modules to learn styles for StyleGANs, Wei~\etal~\cite{wei2021simpleinversion} propose a simple and efficient head, which just consists of an average pooling layer and a fully connected layer. 
Given three different semantic levels of features obtained by the feature pyramid network (FPN)~\cite{lin2017feature}, these three heads produce $\w_{15},\cdots,\w_{18}$, $\w_{10},\cdots,\w_{14}$, and $\w_1,\cdots,\w_9$ from the shallow, medium, and deep features, respectively.
In~\cite{tov2021designing}, Tov~\etal analyze the trade-offs between distortion, perceptual quality, and editability within the StyleGAN latent space.
An encoder is used to control the trade-offs and facilitate downstream image editing.
To improve inversion accuracy, Alaluf~\etal~\cite{alaluf2021restyle} introduce an iterative refinement mechanism for the encoder. 
Instead of directly predicting the latent code of a given real image in a forward pass, at step $t$, the encoder operates on an extended input obtained by concatenating the given image $\x$ with the predicted image: $\Delta_t=E(\x,y_t)$, where $y_t=G(\w_t)$. 
The latent code at step $t+1$ is then updated as $\w_{t+1}=\Delta_t+\w_t$.
The initialized values of $\w_0$ and $y_0$ are set as the average latent code and its corresponding image, respectively.

Although some methods\cite{pidhorskyi2020adversarial,donahue2016adversarial,dumoulin2016adversarially,lang2021explaining} use additive encoder networks to learn the inverse mapping of GANs, we do not categorize them as GAN inversion since their goals are to \emph{jointly train} the encoder with both the generator and the discriminator, instead of determining the latent space of a trained GAN model.

\subsubsection{Optimization-based GAN Inversion}
\label{sec:optimization-based}

Existing optimization-based GAN inversion methods typically reconstruct a target image by optimizing the latent vector
\begin{equation}
\z^* = \underset{\z}{\arg\min}\, \ell(x, G(\z; \theta)),
\label{eqn:opt}
\end{equation}
where $x$ is the target image and $G$ is a GAN generator parameterized by $\theta$.

It is critical to choose the optimizer since a good optimizer helps alleviate the local minima problem.
There are two types of optimizers: gradient-based (ADAM~\cite{kingma2014adam}, L-BFGS~\cite{liu1989limited}, Hamiltonian Monte Carlo (HMC)~\cite{duane1987hybrid}), and gradient-free (covariance matrix adaptation (CMA)~\cite{hansen2001cma}) methods.
Optimization-based GAN inversion methods use different optimizers. 
For example, ADAM~\cite{kingma2014adam} is used in the Image2StyleGAN~\cite{abdal2019image2stylegan}, and L-BFGS is used by Zhu~\etal~\cite{zhu2016generative}.
Huh~\etal~\cite{huh2020transforming} systematically experiment with different choices of both gradient-based and gradient-free optimizers and find that CMA and its variant BasinCMA perform the best for optimizing the latent vector when inverting images in challenging datasets (\eg~LSUN Cars~\cite{yu2015lsun}) to the latent space of StyleGAN2~\cite{karras2020analyzing}.

Another important issue for optimization-based GAN inversion is the initialization of latent code. 
Since Equation~\eqref{eqn:def} is highly nonconvex, the reconstruction quality strongly relies on a good initialization of $\z$ (sometimes $\w$ for StyleGAN~\cite{karras2019style}).
Experiments show that different initial values lead to a significant perceptual difference in generated images~\cite{radford2016dcgan,brock2018large,karras2017progressive,karras2019style}. 
An intuitive solution is to start with several random initial values and obtain the best result with minimal cost. 
Image2StyleGAN~\cite{abdal2019image2stylegan} studies two initialization choices, one based on random selection and the other based on mean latent code $\overline{\w}$. %
However, a prohibitively large number of random initial values may be tested before obtaining a stable reconstruction~\cite{zhu2016generative}, which makes real-time processing impossible. 
Thus, some~\cite{zhu2016generative,tewari2020stylerig} instead train a deep neural network to minimize~\eqref{eqn:def} directly, as introduced in Section~\ref{sec:learning-based}.
Some~\cite{zhu2016generative,bau2019inverting} propose using an encoder to provide better initialization for optimization, which is discussed in Section~\ref{sec:hybrid}.

We note that the optimization-based methods~\cite{creswell2018inverting,abdal2019image2stylegan,abdal2020image2stylegan2} typically require an expensive iterative process in terms of both memory and runtime, as they have to be applied to each latent code independently. 

\subsubsection{Hybrid GAN Inversion}
\label{sec:hybrid}

The hybrid methods~\cite{zhu2016generative,bau2019seeing,bau2019inverting,zhu2020indomain} exploit the advantages of both approaches discussed above. 
As one of the pioneering works in this field, Zhu~\etal~\cite{zhu2016generative} propose a framework that first predicts $\z$ of a given real photo $x$ by training a separate encoder $E(x; \theta_E)$, which then uses the obtained $\z$ as the initialization for optimization.
The learned predictive model serves as a fast bottom-up initialization for the nonconvex optimization problem~\eqref{eqn:def}.

Subsequent studies follow this framework and have proposed several variants.
For example, to invert $G$, Bau~\etal~\cite{bau2019inverting} begin by training a network $E$ to obtain a suitable initialization of the latent code $\z_{0}=E(x)$ and its intermediate representation $\rr_{0}=g_{n}(\cdots(g_{1}(\z_{0})))$, where $g_{n}(\cdots(g_{1}(\cdot)))$ in a layerwise representation of $G(\cdot)$.
This method then uses $\rr_{0}$ to initialize a search for $\rr^{*}$ to obtain a reconstruction $x^{\prime}=G(\rr^{*})$ close to the target $x$ 
(see Section~\ref{sec:layerwise} for more details).
Zhu~\etal~\cite{zhu2020indomain} show that in most existing methods, generator $G$ does not provide its domain knowledge to guide the training of encoder $E$ since the gradients from $G(\cdot)$ are not taken into account at all. 
To fix it, a domain-specific GAN inversion approach is developed, which both reconstructs the input image and ensures that the inverted code is meaningful for semantic editing (see Section~\ref{sec:semantic-aware} for more details of this method).
In contrast to previous methods, Roich~\etal~\cite{roich2021pivotal} develop a generator-tuning technique. 
Using an initial latent code as the pivot, they lightly tune the pretrained generator so that the input image can be faithfully reconstructed. 
This process is referred to as pivotal tuning, which helps map an out-of-domain image to an in-domain latent code~\cite{zhu2020indomain} faithfully.
Alaluf~\etal~\cite{alaluf2021hyperstyle} further introduce a hypernetwork~\cite{ha2016hypernetworks} that learns to refine the generator weights with respect to a given input image. 
The hypernetwork is composed of a lightweight feature extractor and a set of refinement blocks.

\subsection{Properties of GAN Inversion Methods}
\label{sec:characteristics}

In this section, we discuss the important properties of GAN inversion methods, \ie, \textit{having supported resolution}, \textit{being semantic-aware}, \textit{being layerwise}, and \textit{having out-of-distribution generalizability}. 

\subsubsection{Supported Resolution}
\label{sec:resolution}

\figmodel

The image resolution that a GAN inversion method can support is mainly determined by the capacity of generators and inversion mechanisms. 
Zhu~\etal~\cite{zhu2016generative} use GCGANs trained on several datasets with images of $64 \times 64$ pixels, and 
Bau~\etal~\cite{bau2019gandissect,bau2019ganpaint} adopt PGGANs~\cite{karras2017progressive} trained with images of size $256 \times 256$ pixels from Lsun~\cite{yu2015lsun}.
However, some methods cannot fully leverage the pretrained GAN model. 
Zhu~\etal~\cite{zhu2020indomain} propose an encoder to map the given images to the latent space of StyleGAN. 
This method (Fig.~\ref{fig:modelcompare} (a)) performs well for images of $256 \times 256$ pixels but does not scale up well to images of $1024 \times 1024$ pixels due to the high computational cost 
(where 1/n in the figure means semantic feature maps of 1/n original input resolution).
Conversely, the pSp method proposed by Richardson~\etal~\cite{richardson2020encoding} (Fig.~\ref{fig:modelcompare} (b)) can synthesize images of $1024 \times 1024$ pixels, regardless of input image size, since the 18 map2style modules they proposed are used to predict 18 single-layer latent codes separately.
Wei~\etal~\cite{wei2021simpleinversion} propose a similar model 
but with a lightweight encoder.
Similar to~\cite{richardson2020encoding}, features from three semantic levels are used to predict different parts of the latent codes.
Nevertheless, this model predicts 9, 5, and 4 layers of latent codes from each semantic level, as shown in Fig.~\ref{fig:modelcompare} (c).
Recent applications such as face swapping on megapixels~\cite{zhu2021megafs,bai2021identity} and infinite-resolution image synthesis~\cite{lin2021infinity} are developed as image inversion methods that can support high-resolution image editing.

\subsubsection{Semantic Awareness}
\label{sec:semantic-aware}

GAN inversion methods with semantic-aware properties can perform image reconstruction at the pixel level and align the inverted code with the knowledge that emerge in the latent space. 
Semantic-aware latent codes can better support image editing by reusing the rich knowledge encoded in the GAN models.
The existing approaches typically randomly sample a collection of latent codes $\z$ and feed them into $G(\cdot)$ to obtain the corresponding synthesis $x^{\prime}$. 
The encoder $E(\cdot)$ is then trained by
\begin{equation}
\min_{\Theta_{E}} \mathcal{L}_{E}=\|\z-E(G(\z))\|_{2},
\end{equation}
where $\|\cdot\|_{2}$ denotes the $l_{2}$ distance, and $\Theta_{E}$ represents the parameters of the encoder $E(\cdot)$.
Collins~\etal~\cite{collins2020uncovering} use a latent object representation to synthesize images with different styles and reduce artifacts.
However, the supervision by only reconstructing $\z$ (or equivalently, the synthesized data) is not sufficient to train an accurate encoder. 

To alleviate this issue, Zhu~\etal~\cite{zhu2020indomain} propose a domain-specific GAN inversion approach to recover the input real image at both the pixel and semantic levels.
This method first trains a domain-guided encoder $E$ to map the image space to the latent space such that all codes produced by the encoder are in-domain latent codes.
The encoder $E$ is trained to recover the real images, instead of being trained with synthesized data to recover the latent code. 
Then, they perform the instance-level domain-regularized optimization by involving this well-trained $E$ as a regularization term to fine-tune the latent code in the semantic domain during $\mathbf{z}$ optimization.
Such optimization helps better reconstruct the pixel values without affecting the semantic property of the inverted code.
The training process is formulated as
\begin{equation} 
\begin{aligned}
\min_{\Theta_{E}} \mathcal{L}_{E}=\|x-G(E(x))\|_{2} 
&+\lambda_1\|F(x)-F(G(E(x)))\|_{2} \\ 
&-\lambda_2 {\E}[D(G(E(x)))],
\end{aligned}
\end{equation} 
where $F(\cdot)$ represents the VGG feature extraction, $\E[D(\cdot)]$ is the discriminator loss, and $\lambda_1$ and $\lambda_2$ are the perceptual and discriminator loss weights, respectively.
The inverted code from the proposed domain-guided encoder can well reconstruct the input image based on the pretrained generator and ensure the code itself to be semantically meaningful. However, the code still needs refinement to better fit the individual target image at the pixel values.
Based on the domain-guided encoder, they design a domain-regularized optimization with two modules:
(i) the output of the domain-guided encoder is used as a starting point to avoid a local minimum and also shorten the optimization process, and (ii) a domain-guided encoder is used to regularize the latent code within the semantic domain of the generator. 
The objective function is
\begin{equation}
\begin{aligned}
\z^{*}=\underset{\z}{\arg \min }\|x-G(\z)\|_{2} &+\lambda_1^{\prime}\|F(x)-F(G(\z))\|_{2} \\
&+\lambda_2^{\prime}\|\z-E(G(\z))\|_{2},
\end{aligned} 
\end{equation}
where $x$ is the target image to invert, and $\lambda_1^{\prime}$ and $\lambda_2^{\prime}$ are the loss weights corresponding to the perceptual loss and the encoder regularizer, respectively.

\subsubsection{Layerwise}
\label{sec:layerwise}

When the number of layers is large, it is not feasible to determine the generator for the full inversion problem defined by Equation (1). 
Some recent approaches~\cite{bau2019seeing,lei2019inverting} are developed to solve a tractable subproblem by decomposing the generator $G$ into layers:
\begin{equation}
G=G_{f}(g_{n}(\cdots((g_{1}(\z)))),
\end{equation}
where $g_{1}, \ldots, g_{n}$ are the early layers of $G$, and $G_{f}$ constructs all the later layers of $G$.

The simplest layerwise GAN inversion is based on one layer.
Lei~\etal~\cite{lei2019inverting} consider a one-layer model in the form of $G=g(\z)=\text{ReLU}(\W\z+\bb)$.
When the problem is realizable,
to find a feasible $\z$ such that $x= G(\z)$, one could invert the function by solving a linear programming problem:
\begin{eqnarray}
{\w_i}^\top \z + b_i = x_i, &\forall i \text{ s.t. } x_i>0, \nonumber\\
{\w_i}^\top \z + b_i\leq 0, &\forall i \text{ s.t. } x_i= 0.
\label{eqn:single_layer}
\end{eqnarray}
The solution set of~\eqref{eqn:single_layer} is convex and forms a polytope. 
However, it possibly includes uncountable feasible points~\cite{lei2019inverting}, which makes it unclear how to conduct layerwise inversion.
Several methods make additional assumptions to generalize the above result to deeper neural networks. 
Lei~\etal~\cite{lei2019inverting} assume that the input signal is corrupted by bounded noise in terms of $\ell_1$ or $\ell_{\infty}$ and propose an inversion scheme for generative models using linear programs layer by layer. 
The analysis for an assuredly stable inversion is restricted to cases where the following hold: 
(1) the weights of the network should be Gaussian \iid~variables; 
(2) each layer should be expanded by a constant factor; and
(3) the last activation function should be 
ReLU~\cite{nair2010rectified} or leaky ReLU~\cite{maas2013rectifier}.
However, these assumptions often do not hold in practice. 

To invert complex state-of-the-art GANs, Bau~\etal~\cite{bau2019seeing} propose solving the easier problem of inverting the final layers $G_f$:
\begin{equation}
x^{\prime} = G_f(\rr^{*}),
\label{eqn:ganseeing}
\end{equation}
where $\rr^{*} = \underset{\rr}{\arg\min} \ell(G_f(\rr), x)$, $\rr$ is an intermediate representation, and $\ell$ is a distance metric in the image feature space. 
They solve the inversion problem~\eqref{eqn:def} in a two-step hybrid GAN inversion framework: first constructing a neural network $E$ that approximately inverts the entire $G$ and computes an estimate $\z_0 = E(x)$ and subsequently solving an optimization problem to identify $\rr^* \approx \rr_0 = g_n(\cdots(g_1(\z_0)))$ that generates a reconstructed image $G_f(\rr^*)$ to closely recover $x$.
For each layer $g_i \in \{g_1,...,g_n, G_f\}$, a small network $e_i$ is first trained to invert $g_i$. 
That is, when defining $\rr_i = g_i(\rr_{i-1})$, the goal is to learn a network, $e_i$, that approximates the computation $\rr_{i-1} \approx e_i(\rr_{i})$ and ensures the predictions of the network $e_i$ to well preserve the output of layer $g_i$, \ie, $\mathbf{r_{i}} \approx g_i(e_i(\rr_{i}))$.
As such, $e_i$ is trained to minimize both left- and right-inversion losses:
\begin{equation}
\begin{aligned}
\mathcal{L}_{\text{L}} & = \E_{\z}[||\rr_{i-1}- e(g_i(\rr_{i-1}))||_1], \\
\mathcal{L}_{\text{R}} & = \E_{\z}[||\rr_i - g_i(e(\rr_i))||_1],\\
e_i & = \underset{e}{\arg\min}\quad \mathcal{L}_{\text{L}} + \lambda_{\text{R}}\; \mathcal{L}_{\text{R}},
\end{aligned}
\end{equation}
where $||\cdot||_1$ denotes an $\mathcal{L}_1$ loss, and $\lambda_{\text{R}}$ is set as 0.01 to emphasize the reconstruction of $\rr_{i-1}$.
To focus on training near the manifold of representations produced by the generator, this method uses sample $\z$ and layers $g_i$ to compute samples of $\rr_{i-1}$ and $\rr_i$ such that $\rr_{i-1} = g_{i-1}(\cdots g_1(\z))$.
Once all the layers are inverted, an inversion network for all of $G$ can be composed as follows:
\begin{equation}
{E}^{*}= e_1(e_2(\cdots(e_n(e_f(x))))).
\end{equation}
The results can be further improved by fine-tuning the composed network $E^*$ to invert $G$ jointly as a whole and obtain the final result $E$.

For StyleGANs~\cite{karras2019style,karras2020analyzing,Karras2020ada}, the intermediate latent vector $\w \in \mathcal{W}^{+}$ or $\s \in \mathcal{S}$ is different across layers and is fed into the corresponding layer of the generator via AdaIN~\cite{huang2017adain} or affine transformations~\cite{wu2020stylespace}.
Therefore, inverting images into $\mathcal{W}^{+}$ or $\mathcal{S}$ space can be seen as being layerwise.

\subsubsection{Out-of-Distribution Generalizability}
\label{sec:ood}

GAN inversion methods can support inverting the images, especially any given real images that are not generated by the same process of the training data. 
We refer to this ability as out-of-distribution generalizability~\cite{ren2019likelihood,hendrycks2016baseline,lee2018simple}.
Specifically, given a StyleGAN pretrained on the FFHQ dataset, this property is closely related to the following two aspects: 1) to generate face images with all combinations of facial attributes, even if some combinations do not exist in the training dataset; 2) to handle the images different to the samples of the training set, such as corrupted images, caricatures, or black and white photos.
This property is a prerequisite for GAN inversion methods to edit a wider range of images.
Out-of-distribution generalizability has been demonstrated in many GAN inversion methods.
Zhu~\etal~\cite{zhu2020indomain} propose a domain-specific GAN inversion approach to recover the input image at both the pixel and semantic levels.
Although trained only with the FFHQ dataset, their model can generalize to not only real face images from multiple face datasets~\cite{chelnokova2014rewards,courset2018caucasian,yi2019apdrawinggan} but also paintings, caricatures, and black and white photos collected from the Internet.
Kang~\etal~\cite{kang2021gan} propose a method to invert out-of-range images. 
Taking facial images as an example, out-of-range images could be the images with extreme poses or the corrupted images, which previous methods often fail to handle.
Being able to invert out-of-range images allows GAN inversion methods to be applied to wider domains rather than limited settings.
Some methods~\cite{abdal2020image2stylegan2,chai2021using} explore the potential of inverting an image into a desired latent code just given a degraded or partial observation. 
In addition to images, recent methods also show out-of-distribution generalization ability for other modalities, \ie, sketch~\cite{richardson2020encoding,wei2021simpleinversion} and text~\cite{xia2021open,patashnik2021styleclip}.

The out-of-distribution generalizability of GAN inversion facilitates open-world image manipulation when combined with the latent code-based editing methods (see Section 4.4)~\cite{han2021IALS,voynov2020latent,nurit2021steerability,shen2021closedform}.
One notable drawback is that inverting images that contain unseen attributes can easily lead to unexpected results as they lie outside the domain of the pretrained image generators. 
This limits extending GAN inversion to broader applications such as image synthesis guided by uncommon textual descriptions~\cite{xia2021open}.
Some recent approaches aim to alleviate this issue by transferring the GANs pretrained on one image domain to a new one, guided by 
certain references or semantics from
one or few target images~\cite{yang2021genda} (few-shot and one-shot), pretrained language-image models~\cite{gal2021clip} (zero-shot), or both~\cite{zhu2021mind}.

\subsection{Latent Space Navigation}
\label{sec:navigation}

GAN inversion is not the end goal. The reason that we invert a real image into the latent space of a trained GAN model is that it allows us to manipulate the image by varying the inverted code in the latent space for a certain attribute.
This technique is usually known as latent space navigation or traversals~\cite{zhuang2021enjoy,cherepkov2020navigating}, GAN steerability~\cite{jahanian2020steerability,nurit2021steerability}, or latent code manipulation~\cite{shen2020interpreting}.
Although often regarded as an independent research field, it becomes an indispensable application of the GAN inversion~\cite{patashnik2021styleclip,alaluf2021only}.
Many inversion methods~\cite{tov2021designing,alaluf2021restyle} also explore the efficient discovery of a desired latent code.
Section~\ref{sec:space} has introduced different latent spaces. 
This section introduces discovering interpretable and disentangled directions in the latent spaces of GANs.

\subsubsection{Discovering Interpretable Directions}
\label{sec:interpretable}
Some methods support discovering interpretable directions in the latent space, \ie, controlling the generation process by varying the latent codes $\z$ in the desired directions $\n$ with step $\alpha$, which is considered as the vector arithmetic $\z^{\prime}=\z+\alpha\n$.
Such directions can be identified through supervised, unsupervised, or self-supervised manners. Recent methods have also been proposed to directly compute the interpretable directions in closed form from the pretrained models without any kind of training or optimization.

\vspace{1mm}
\noindent\textbf{Supervised Setting.} 
Existing supervised learning-based approaches typically randomly sample a large amount of latent codes, synthesize a collection of corresponding images, and annotate them with some predefined labels by introducing a pretrained classifier (\eg, predicting face attributes or light directions)~\cite{goetschalckx2019ganalyze,shen2020interpreting,abdal2020styleflow,jahanian2020steerability} or extracting statistical image information (\eg, color variations)~\cite{plumerault2020control}.
For example, to interpret the face representation learned by GANs, Shen~\etal~\cite{shen2020interpreting} employ some off-the-shelf classifiers to learn a hyperplane in the latent space serving as the separation boundary and predict semantic scores for synthesized images.
Abdal~\etal~\cite{abdal2020styleflow} learn a semantic mapping between the $\mathcal{Z}$ space and the $\mathcal{W}$ space by using continuous normalizing flows (CNF).
Both methods rely on the availability of attributes (typically obtained by a face classifier network), which might be difficult to obtain for new datasets and could require manual labeling effort.

\fignoninference

\vspace{1mm}
\noindent\textbf{Unsupervised Setting.} 
The supervised setting would introduce bias into the experiment since the sampled codes and synthesized images used as supervision are different in each sampling and may lead to different discoveries of interpretable directions~\cite{shen2021closedform}. 
It also severely restricts a range of directions that existing approaches can discover, especially when the labels are missing. 
Furthermore, the individual controls discovered by these methods are typically entangled, affecting multiple attributes, and are often nonlocal.
Thus, some methods~\cite{voynov2020latent,lu2020discovery,eric2020GANSpace,cherepkov2020navigating} aim to discover interpretable directions in the latent space in an unsupervised manner, \ie, without the requirement of paired data.
For example, Härkönen~\etal~\cite{eric2020GANSpace} create interpretable controls for image synthesis by identifying important latent directions based on PCA applied in the latent or feature space. The obtained principal components correspond to certain attributes, and the selective application of the principal components allows for the control of many image attributes.
This method is considered as ``unsupervised'' since the directions can be discovered by PCA without using any labels.
Manual intervention and supervision are required to annotate these directions to the target operations and to which layers they should be applied to.
In contrast, Jahanian~\etal~\cite{jahanian2020steerability} optimize trajectories (both linear and nonlinear) in a self-supervised manner.
Taking the linear walk $\ww$ as an example, given an inverted source image $G(\z)$, they learn $\ww$ as
\begin{equation}
\w^* = \underset{\w}{\arg\min} {\E}_{\z,\alpha} [\mathcal{L} ( G(\z\!+\!\alpha \w), \texttt{edit}(G(\z), \alpha))],
\label{eq:optimal_w}
\end{equation}
where $\mathcal{L}$ measures the distance between the generated image $G(\z+\alpha \ww)$ after taking an $\alpha$-step in the latent direction and the target image \texttt{edit}($G(\z), \alpha$). 
This method is considered as ``self-supervised'' because the target image ($G(\z), \alpha$) could be derived from the source image $G(\z)$.

\vspace{1mm}
\noindent\textbf{Closed-form Solution.}
A few methods~\cite{shen2021closedform,nurit2021steerability,wei2021orthogonal,zhu2021lowrankgan} recent show that interpretable directions for image synthesis can be directly obtained in closed forms without training or optimization. 
Shen~\etal~\cite{shen2021closedform} propose a semantic factorization method based on the singular value decomposition of the weights of the first layer of a pretrained GAN.
They observe that the semantic transformation of an image, usually denoted by moving the latent code toward a certain direction $\n^{\prime} = \z + \alpha \n$, is actually determined by the latent direction $\n$, which is independent of the sampled code $\z$.
A \textbf{Se}mantics \textbf{Fa}ctorization (SeFa) method is developed to discover the directions $\n$ that can cause a significant change in the output image $\Delta\y$, \ie, $\Delta\y =\y^{\prime}-\y= (\A(\z + \alpha\n) + \bb) - (\A\z + \bb) = \alpha\A\n$, where $\A$ and $\bb$ are the weight and bias of certain layers in $G$, respectively. 
The obtained formula, $\Delta\y = \alpha\A\n$, suggests that the desired editing with direction $\n$ can be achieved by adding the term $\alpha\A\n$ onto the projected code and indicates that the weight parameter $\A$ should contain the essential knowledge of image variations.
The problem of exploring the latent semantics can thus be factorized by solving the following optimization problem:
\begin{equation}
  \n^* = \underset{\{\n\in\R^d:\ \n^T\n = 1\}}{\arg\max} ||\A\n||_2^2.  
  \label{eq:single-optimization}    
\end{equation}
The desired directions $\n^*$, \ie, a closed-form factorization of latent semantics in GANs, should be the eigenvectors of the matrix $\A^T\A$.
In contrast to SeFa~\cite{shen2021closedform},
a method based on orthogonal Jacobian regularization is 
applied to multiple layers of the generator to determine interpretable directions for image synthesis~\cite{wei2021orthogonal}.

\subsubsection{Discovering Disentangled Directions}
\label{sec:disentangled}

When several attributes are involved, editing one may affect another since some semantics are not separated.
Some methods aim to tackle multi-attribute image manipulation without interference.
This characteristic is also named multidimensional~\cite{nitzan2020harness} or conditional editing~\cite{shen2020interpreting} in the literature.
The goal is to discover disentangled directions for the desired attributes.
For example, to edit multiple attributes, Shen~\etal~\cite{shen2020interpreting} formulate the inversion-based image manipulation as $x^{\prime}=G(\z^{*}+\alpha \n)$, where $\n$ is a unit normal vector indicating a hyperplane defined by two latent codes $\z_{1}$ and $\z_{1}$.
In this method, $k$ attributes $\{\z_{1}, \cdots, \z_{k}\}$ can form $m$ (where $m \leq k(k-1)/2$) hyperplanes $\{\n_{1}, \cdots, \n_{m}\}$. 
To edit multiple attributes without interfering with each other, these disentangled directions $\{\n_{1}, \cdots, \n_{m}\}$ should be orthogonal.
If this condition does not hold, then some semantics will correlate with each other, and $\n_{i}^{\top} \n_{j}$ can be used to measure the entanglement between the $i$-th and $j$-th semantics.
In particular, this method uses projection to orthogonalize different vectors. 
As shown in Fig.~\ref{fig:projection}, given two hyperplanes with normal vectors $\n_{1}$ and $\n_{2}$, the goal is to find a projected direction $\n_{1}-(\n_{1}^{\top} \n_{2}) \n_{2}$ such that moving samples along this new direction can change ``attribute one'' without affecting ``attribute two''. 
For the case where multiple attributes are involved, they subtract the projection from the primal direction onto the plane that is constructed by all conditioned directions.
Other GAN inversion methods~\cite{viazovetskyi2020distillation} based on pretrained StyleGAN~\cite{karras2019style} or StyleGAN2~\cite{karras2020analyzing} models can also manipulate multiple attributes due to the stronger separability of $\mathcal{W}$ space than of $\mathcal{Z}$ space.
However, as observed by recent methods~\cite{xia2021tedigan,wu2020stylespace}, some attributes remain entangled in the $\mathcal{W}$ space, leading to some unwanted changes when we manipulate a given image.
Instead of manipulating in the semantic $\mathcal{W}$ space, Wu~\etal~\cite{wu2020stylespace} propose the $\mathcal{S}$ space (style space).
The style code is formed by concatenating the output of all affine layers of the StyleGAN2~\cite{karras2020analyzing} generator.
Experiments show that the $\mathcal{S}$ space can alleviate \textit{spatially entangled changes} and exert precise local modifications.
By intervening the style code $s \in \mathcal{S}$ directly, their method can manipulate different facial attributes along with various semantic directions without affecting others and can achieve fine-grained controls on local translations.

\section{Applications}
\label{sec:applications}

Finding an accurate solution to the inversion problem allows us to match the target image without compromising the editing capabilities in the downstream tasks.
GAN inversion does not require task-specific dense-labeled datasets and can be applied to many tasks such as image manipulation, image interpolation, image restoration, style transfer, novel-view synthesis, and even adversarial defense.
In addition to the common image editing applications, in the last few months, GAN inversion techniques have been widely introduced to many other tasks, such as 3D reconstruction~\cite{pan2020gan2shape,zhang2021unsupervised}, image understanding~\cite{Tritrong2021RepurposeGAN,abdal2021labels4free}, multimodal learning~\cite{xia2021tedigan,xia2021open,patashnik2021styleclip,wang2021cigan}, and medical imaging~\cite{ren2021medical,fetty2020latent,daroach2021high}, which shows its versatility for different tasks and strength to benefit a larger research community.

\subsection{Image Manipulation}
\label{sec:manipulation}

Given an image $x$, we want to edit certain regions by varying its latent codes $\z$ and obtain ${\mathbf{z^{\prime}}}$ of the target image ${x^{\prime}}$ by linearly transforming the latent representation from a trained GAN model $G$. 
This can be formulated in the framework of GAN inversion as the operation of adding a scaled difference vector:
\begin{equation}
x^{\prime}=G(\z^{*}+\alpha \n), 
\end{equation}
where $\n$ is the normal direction corresponding to a particular semantic in the latent space, and $\alpha$ is the step for manipulation. 
In other words, if a latent code is moved in a certain direction, then the semantics contained in the output image should vary accordingly. 
For example, Voynov~\etal~\cite{voynov2020latent} gradually determine the direction corresponding to the background removal or background blur without changing the foreground.
Shen~\etal~\cite{shen2020interpreting} achieve single and multiple facial attribute manipulation by projecting and orthogonalizing different vectors.
Recently, Zhu~\etal~\cite{zhu2020indomain} perform semantic manipulation by either decreasing or increasing the semantic degree. 
Both methods~\cite{shen2020interpreting,zhu2020indomain} use a projection strategy to search for the semantic direction $\n$.

Some methods can perform region-of-interest editing, which allows for the editing of some desired regions in a given image with user manipulation.
Such operations often involve additional tools to select the desired region.
For example, Abdal~\etal~\cite{abdal2019image2stylegan,abdal2020image2stylegan2} analyze the defective image embedding of StyleGAN trained on FFHQ~\cite{karras2019style}, \ie, the embedding of images with masked regions.
The experiments show that the StyleGAN embedding is quite robust to the defects in images, and the embeddings of different facial features are independent of each other~\cite{abdal2019image2stylegan}. 
Based on their observation, they develop a mask-based local manipulation method.
They find a plausible embedding for regions outside the mask and fill in reasonable semantic content in the masked pixels. 
Zhu~\etal~\cite{zhu2020indomain} use their in-domain inversion method for semantic diffusion. This task is to insert the target face into the context and makes them compatible.
Their method can keep the salient features of the target image (\eg, face identity) and adapt to the context information at the same time.

Some methods also can manipulate the image other than the semantics, \eg, geometry, texture, and color. 
For example, \cite{abdal2020styleflow,abdal2019image2stylegan} change pose rotation for face manipulation, while \cite{voynov2020latent} can manipulate geometry (\eg, zoom/shift/rotation), texture (\eg, background blur/add grass/sharpness), and color (\eg, lighting/saturation).

\subsection{Image Generation}
\label{sec:generation}

Several GAN inversion-based methods are proposed for image generation tasks, such as hairstyle transfer~\cite{saha2021LOHO}, few-shot semantic image synthesis~\cite{endo2021fewshotsmis}, and infinite-resolution image synthesis~\cite{lin2021infinity}. 
Saha~\etal~\cite{saha2021LOHO} develop a photorealistic hairstyle transfer method by optimizing the extended latent space and the noise space of StyleGAN2~\cite{karras2020analyzing}.
Endo~\etal~\cite{endo2021fewshotsmis} assume pixels sharing the same semantics have similar StyleGAN features to generate images and corresponding pseudosemantic masks from random noise in the latent space, and use a nearest-neighbor search for synthesis. 
This method integrates an encoder with the fixed StyleGAN generator and trains the encoder with the pseudolabeled data in a supervised fashion to control the generator.
Cheng~\etal~\cite{cheng2021InOut} propose a GAN inversion-based method for image inpainting and outpainting. 
A coordinate-conditioned generator is designed to synthesize patches to be concatenated for a full image.
The latent codes, depending on the joint latent codes and their coordinates, synthesize the images overlapping with the input image.
The optimal latent code for the available input patches is determined in the latent space of the trained patch-based generator during the outpainting stage.
GAN inversion methods can be applied to interactive generation, \ie, starting with strokes drawn by a user and generating natural images that best satisfy the user constraints. 
Zhu~\etal~\cite{zhu2016generative} show that users can employ the brush tools to generate an image from scratch and then continually add more scribbles to refine the result.
Abdal~\etal~\cite{abdal2020image2stylegan2} invert the StyleGAN to perform semantic local edits based on user scribbles. 
With this method, simple scribbles can be converted into photorealistic edits 
by embedding them into certain layers of StyleGAN.
This application is helpful for existing interactive image processing tasks such as sketch-to-image generation~\cite{xia2019sketch,ghosh2019isketchnfill,chenDeepFaceDrawing2020} and sketch-based image retrieval~\cite{eitz2010sketch,dey2019doodle}, which usually require densely labeled datasets.

\subsection{Image Restoration}
\label{sec:restoration}

Suppose that $\hat{x}$ is obtained via $\hat{x}=\phi(x)$ during acquisition, where $x$ is the distortion-free image, and $\phi$ is a degradation transform.
Many image restoration tasks can be regarded as recovering $x$ given $\hat{x}$. 
A common practice is to learn a mapping from $\hat{x}$ to $x$, which often requires task-specific training for different $\phi$.
Alternatively, GAN inversion can employ statistics of $x$ stored in some prior and search in the space of $x$ for an optimal $x$ that best matches $\hat{x}$ by viewing $\hat{x}$ as partial observations of $x$.
For example, Abdal~\etal~\cite{abdal2019image2stylegan,abdal2020image2stylegan2} observe that StyleGAN embedding is quite robust to the defects in images, \eg,~masked regions.
Based on that observation, they propose an inversion-based image inpainting method by embedding the source defective image into the early layers of the $\mathcal{W}^{+}$ space to predict the missing content and into the later layers to maintain color consistency.
Pan~\etal~\cite{pan2020exploiting} claim that a fixed GAN generator is inevitably limited by the distribution of training data and its inversion cannot faithfully reconstruct unseen and complex images.
Thus, they present a relaxed and more practical reconstruction formulation for capturing the statistics of natural images in a trained GAN model as do the prior methods, \ie, the deep generative prior (DGP).
Specifically, they reformulate \eqref{eqn:opt} such that it allows the generator parameters to be fine-tuned on the target image on the fly:
\begin{equation}
\theta^*,\, \z^* = \underset{\theta,\, \z}{\arg\min}\, \ell(\hat{x}, \phi(G(\z; \theta))).
\label{eqn:opt_dgp}
\end{equation}
Their method performs comparable to state-of-the-art methods in terms of colorization~\cite{larsson2016learning}, inpainting~\cite{ulyanov2018deep}, and super-resolution~\cite{shaham2019singan}.
While artifacts sometimes occur in synthesized face images by GAN models~\cite{karras2017progressive,karras2019style}, 
Shen~\etal~\cite{shen2020interpreting} show that the quality information encoded in the latent space can be used for restoration. 
The artifacts generated by PGGAN~\cite{karras2017progressive} can be corrected by moving the latent code toward the positive quality direction that is defined by a separating hyperplane using a linear SVM~\cite{cortes1995support}.

\subsection{Image Interpolation}
\label{sec:interpolation}

With GAN inversion, new results can be interpolated by morphing between corresponding latent vectors of given images.
Given a well-trained GAN generator $G$ and two target images $x_{A}$ and $x_{B}$, morphing between them could naturally be achieved by interpolating between their latent vectors $\z_{A}$ and $\z_{B}$. 
Typically, morphing between $x_{A}$ and $x_{B}$ can be obtained by applying linear interpolation~\cite{xia2020gaze,pan2020exploiting}: 
\begin{equation}
\z=\lambda \z_{A}+(1-\lambda) \z_{B}, \lambda \in(0,1).
\label{eqn:interp}
\end{equation}

Such an operation can be found in ~\cite{nitzan2020harness,abdal2019image2stylegan}.
Moreover, in DGP~\cite{pan2020exploiting}, reconstructing two target images $x_{A}$ and $x_{B}$ would result in two generators $G_{\theta_{A}}$ and $G_{\theta_{B}}$, respectively, and the corresponding latent vectors $\z_{A}$ and $\z_{B}$ since they also fine-tune $G$. 
In this case, morphing between $x_{A}$ and $x_{B}$ can be achieved by linear interpolation of both the latent vectors and the generator parameters: 
\begin{equation}
\begin{aligned}
\z&=\lambda \z_{A}+(1-\lambda) \z_{B},\\ \theta&=\lambda \theta_{A}+(1-\lambda) \theta_{B}, \; \lambda \in(0,1),
\end{aligned}
\label{eqn:interp_dgp}
\end{equation}
and images can be generated with the new $\z$ and $\theta$.

\subsection{3D Reconstruction}
\label{sec:3d}

For 3D data, Pan~\etal~\cite{pan2020gan2shape} and Zhang~\etal~\cite{zhang2021unsupervised} propose 3D shape reconstruction from single images and point cloud completion based on GAN inversion. 
Given an image generated by GAN, starting with an initial ellipsoid 3D object shape, Pan~\etal~\cite{pan2020gan2shape} first render a number of unnatural images with various randomly sampled viewpoints and lighting conditions (called pseudosamples). By reconstructing them with the GAN, these pseudosamples could guide the original image toward the sampled viewpoints and lighting conditions in the GAN manifold, producing a number of natural-looking images (called projected samples). These projected samples could be adopted as the ground truth of the differentiable rendering process to refine the prior 3D shape.
Instead of using existing 2D GANs trained on images, Zhang~\etal~\cite{zhang2021unsupervised} first train a generator $G$ on 3D shapes in the form of point clouds. 
Latent codes are used by the pretrained generator to produce complete shapes. 
Given a partial shape, they look for a target latent vector $\mathbf{z}$ and fine-tune the parameters $\theta$ of $G$ that best reconstruct the complete shape via gradient descent.

\subsection{Image Understanding}
\label{sec:understanding}

A few methods exploit the representations of trained GAN models and leverage these representations for semantic segmentation and alpha matting~\cite{Tritrong2021RepurposeGAN,abdal2021labels4free}.
Tritrong~\etal~\cite{Tritrong2021RepurposeGAN} first embed an image into the latent space for the latent $z$ and feed it into the generator with multiple activation maps.
These maps are upsampled and concatenated along the channel dimension to form the desired representation. 
A segmentation module is trained with a few manually annotated images and  extracted representations.  
During inference, the representation is extracted from a test image and fed into the segmenter to obtain a segmentation map.
In~\cite{abdal2021labels4free},  two pretrained generators, an alpha network and a discriminator are used for the matting task. 
One generator $\mathcal{G}(\z)$ is responsible for generating foreground images, and the other generator $\mathcal{G}_{\mathrm{bg}}\left(\z^{\prime}\right)$ attends to the background. 
The alpha network is used to predict a mask $\mathcal{A}(\z) \odot \mathcal{G}(\z)$ for image matting.
The composite image can be obtained by mixing background and foreground using $\mathcal{A}(\z) \odot \mathcal{G}(\z) + (1-\mathcal{A}(\z)) \odot \mathcal{G}_{\mathrm{bg}}\left(\z^{\prime}\right)$ that the discriminator $\mathcal{D}$ cannot distinguish from the real images.
During training, the two generators are frozen, and only the alpha network and the discriminator are trained by adversarial learning.

\subsection{Multimodal Learning}
\label{sec:multimodality}

For multimodal learning, several recent studies have focused on language-driven image generation and manipulation using StyleGAN. 
Xia~\etal~\cite{xia2021tedigan} propose a novel unified framework for both text-to-image generation and text-guided image manipulation tasks by training an encoder to map texts into the latent space of StyleGAN and perform style-mixing to produce diverse results.
In~\cite{wang2021cigan}, Wang~\etal~propose a similar idea but introduce the cycle-consistency training during inversion to learn more robust and consistent inverted latent codes.
On the other hand, a few methods~\cite{xia2021open,patashnik2021styleclip} first obtain the latent code of a given image and find the target latent code of desired attributes with the guidance of some powerful pretrained language models, \eg,~CLIP~\cite{radford2021learning} or ALIGN~\cite{jia2021scaling}. 
Logacheva~\etal~\cite{logacheva2020video} present a generative model for landscape animation videos based on StyleGAN inversion.
Lee~\etal~\cite{lee2021sound} propose a sound-guided image editing framework. 
They train an audio encoder to encode sounds into a multimodal latent space, where audio representations are aligned with text-image representations to guide image manipulation.

\subsection{Medical Imaging}
\label{sec:medical}

GAN inversion techniques have been recently introduced to medical applications~\cite{yi2019generative}. 
These methods~\cite{fetty2020latent,ren2021medical} are used for data augmentation, where publicly available medical datasets are often outdated, limited, or inadequately annotated.
Typically, these methods train the GAN models on domain-specific medical image datasets, \eg,~Computed Tomography (CT) or Magnetic Resonance (MR), and use existing GAN inversion methods for inversion and manipulation.
Fetty~\etal~\cite{fetty2020latent} present a method based on the StyleGAN model~\cite{abdal2019image2stylegan} in which  CT or MR images with desired attributes can be synthesized by traversing points in the latent space (see Section~\ref{sec:navigation}) or style mixing~\cite{karras2019style}.
To synthesize medical images with desired attributes, Ren~\etal~\cite{ren2021medical} use the domain-specific GAN inversion technique~\cite{zhu2020indomain} to generate mammograms with desired shape and texture for psychophysical experiments.
Overall, these methods based on GAN inversion achieve better interpretability and controllability in medical image synthesis.  

\section{Challenges and Future Directions}
\label{sec:outlook}

\noindent\textbf{Theoretical Understanding.} 
While significant effort has been made on applying GAN inversion to image editing applications, much less attention is paid to a better theoretic understanding of the latent space.
Nonlinear structure in data can be represented compactly, and the induced geometry necessitates the use of nonlinear statistical tools~\cite{kuhnel2018latent}, Riemannian manifolds, and locally linear methods. 
Well-established theories in related areas can facilitate the theoretical understanding from different perspectives.
Some recent methods~\cite{choi2021not,zhu2021lowrankgan} treat the latent space as the manifold structure, which involves different concepts and metrics.

\vspace{1mm}
\noindent\textbf{Inversion Type.} 
In addition to GAN inversion, some methods have been developed to invert generative models based on the encoder-decoder architecture.
The IIN method~\cite{esser2020invertible} learns invertible disentangled interpretations of variational autoencoders (VAEs)~\cite{kingma2013auto}. 
Zhu~\etal~\cite{zhu2019disentangled} develop the latently invertible autoencoder to learn a disentangled representation of face images from which contents can be edited based on attributes. 
The LaDDer approach~\cite{Lin2020LaDDer} uses a meta-embedding based on a generative prior (including an additive VAE and a mixture of hyperpriors) to project the latent space of a well-trained VAE to a lower-dimensional latent space, where multiple VAE models are used to form a hierarchical representation.
It is beneficial to explore combining GAN inversion and encoder-decoder inversion so that we can exploit the best of both worlds. 

\vspace{1mm}
\noindent\textbf{Domain Generalization.}
As discussed in Section~\ref{sec:applications}, GAN inversion proves to be effective in cross-domain applications such as style transfer and image restoration, which indicates that pretrained models have learned domain-agnostic features. 
The images from different domains can be inverted into the same latent space from which effective metrics can be derived. 
Multitask methods have been developed to collaboratively exploit visual cues, such as image restoration and image segmentation~\cite{xia2019adverse} or semantic segmentation and depth estimation~\cite{nekrasov2019joint,zhan2019joint}, within the GAN framework.
It is challenging but worthwhile to develop effective and consistent methods to invert the intermediate shared representations so that we can tackle different vision tasks under a unified framework. 

\vspace{1mm}
\noindent\textbf{Implicit Representation.}
Some methods~\cite{abdal2020styleflow,voynov2020latent} based on pretrained GANs can manipulate geometry (\eg, zoom, shift, and rotate), texture (\eg, background blur and sharpness) and color (\eg, lighting and saturation).
This ability indicates the GAN models pretrained on large-scale datasets have learned some physical information from real-world scenes.
Implicit neural representation learning~\cite{chen2019learning,tucker2020single,rajeswar2020pix2shape}, a recent trend in the 3D computer vision, learns implicit functions for 3D shapes or scenes and enables control of scene properties such as illumination, camera parameters, pose, geometry, appearance, and semantic structure.
It has been used for volumetric performance capture~\cite{chen2020free,liu2020neural,lombardi2019neural}, novel-view synthesis~\cite{martin2020nerf,martin2020nerf}, face shape generation~\cite{wu2020unsupervised}, object modeling~\cite{nguyen2020blockgan}, and human reconstruction~\cite{zheng2020pamir,bhatnagar2020combining,he2020geo,saito2020pifuhd}.
The recent StyleRig method~\cite{tewari2020stylerig} is trained to align the parameters of the 3D morphable model (3DMM)~\cite{egger20203d} with the input of StyleGAN~\cite{karras2019style}.
It opens an interesting research direction to invert such implicit representations of a pretrained GAN for 3D reconstruction, \eg, using StyleGAN~\cite{karras2019style} for human face modeling or time-lapse video generation.

\vspace{1mm}
\noindent\textbf{Precise Control.} 
GAN inversion can be used to find directions for image manipulation while preserving the identity and other attributes~\cite{abdal2020styleflow,shen2020interpreting}.
However, some tuning is needed to achieve the desired granularity of precise control at a fine-grained level, \eg, gaze redirection~\cite{he2019gaze,xia2020gaze}, relighting~\cite{zhou2019deep,zhang2020portrait,sun2019single}, and continuous view control~\cite{chen2019monocular}.
These tasks require precise control, \ie, $1^{\circ}$ of camera view or gaze direction. 
Current GAN inversion methods are incapable of handling the tasks. Thus more efforts are needed, such as creating more disentangled latent spaces and discovering more interpretable directions.

\vspace{1mm}
\noindent\textbf{Multimodal Inversion.}
The existing GAN inversion methods primarily focus on images.
However, recent advances in generative models are beyond the image domain, such as the GPT-3 language model~\cite{brown2020gpt3} and WaveNet~\cite{oord2016wavenet} for audio synthesis. 
Trained on diverse large-scale datasets, these sophisticated deep neural networks prove to be capable of representing an extensive range of different contents, styles, sentiments, and topics.
Applying GAN inversion techniques on these different modalities could provide a novel perspective for tasks such as language style transfer. 
Furthermore, many GAN models are developed for multimodality generation or translation~\cite{li2019control,jia2018speaker,prajwal2020speech}. 
It is a promising direction to invert such GAN models as multimodal representations to create novel kinds of content, behavior, and interaction.

\vspace{1mm}
\noindent\textbf{Evaluation Metrics.}
New perceptual quality metrics, which can better evaluate photorealistic and diverse images or identity consistent with the original image, remain to be explored.
Current evaluations mostly concentrate on measuring photorealism or if the distribution of generated images is consistent with the real images with regard to classification~\cite{bau2019seeing} or segmentation~\cite{voynov2020latent} accuracy using models trained for real images. 
However, there is still a lack of effective assessment tools to evaluate the difference between the predicted results and the expected outcome or to measure the inverted latent codes more directly.

\section{Conclusion}
\label{sec:conclusion}
Deep generative models such as GANs learn the underlying variation factors of the training data through the weak supervision of image generation. Discovering and steering the interpretable latent representations in image generation facilitate a wide range of image editing applications.
This paper presents a comprehensive survey of GAN inversion methods with an emphasis on algorithms and applications. 
We summarize the important properties of GAN latent spaces and models and then introduce four kinds of GAN inversion methods and their key properties. 
We then go through several fascinating applications of GAN inversion, including image manipulation, image generation, image restoration, and recent applications beyond image processing.
We finally discuss the challenges and the future directions of GAN inversion.%

\bibliographystyle{IEEEtran}
\bibliography{IEEEfull}

\ifCLASSOPTIONcaptionsoff
 \newpage
\fi

\end{document}

%% file: commands/newcommand.tex
\newcommand{\eg}{\textit{e.g.}}
\newcommand{\ie}{\textit{i.e.}}
\newcommand{\etal}{\textit{et al.}}

\newcommand{\iid}{\textit{i.i.d.}}

\newcommand{\s}{\mathbf{s}}
\newcommand{\z}{\mathbf{z}}
\newcommand{\x}{\mathbf{x}} 
\newcommand{\y}{\mathbf{y}}

\newcommand{\w}{\mathbf{w}}
\newcommand{\n}{\mathbf{n}} 
\newcommand{\ww}{\boldsymbol{w}}
\newcommand{\bb}{\mathbf{b}}
\newcommand{\rr}{\mathbf{r}}
\newcommand{\A}{\mathbf{A}} 
\newcommand{\W}{\mathbf{W}}

\newcommand{\R}{\mathbb{R}} 
\newcommand{\E}{\mathbb{E}}

\newcommand{\yes}{\checkmark}

\newcommand{\FINAL}[2][]{#2} %
\newcommand{\website}{\FINAL{\texttt{\small{\href{https://github.com/weihaox/awesome-gan-inversion}{github}}}}}

%% file: commands/figure.tex
\newcommand{\figoverview}{
\begin{figure}[tbp]
\begin{center}
\includegraphics[width=0.9\linewidth]{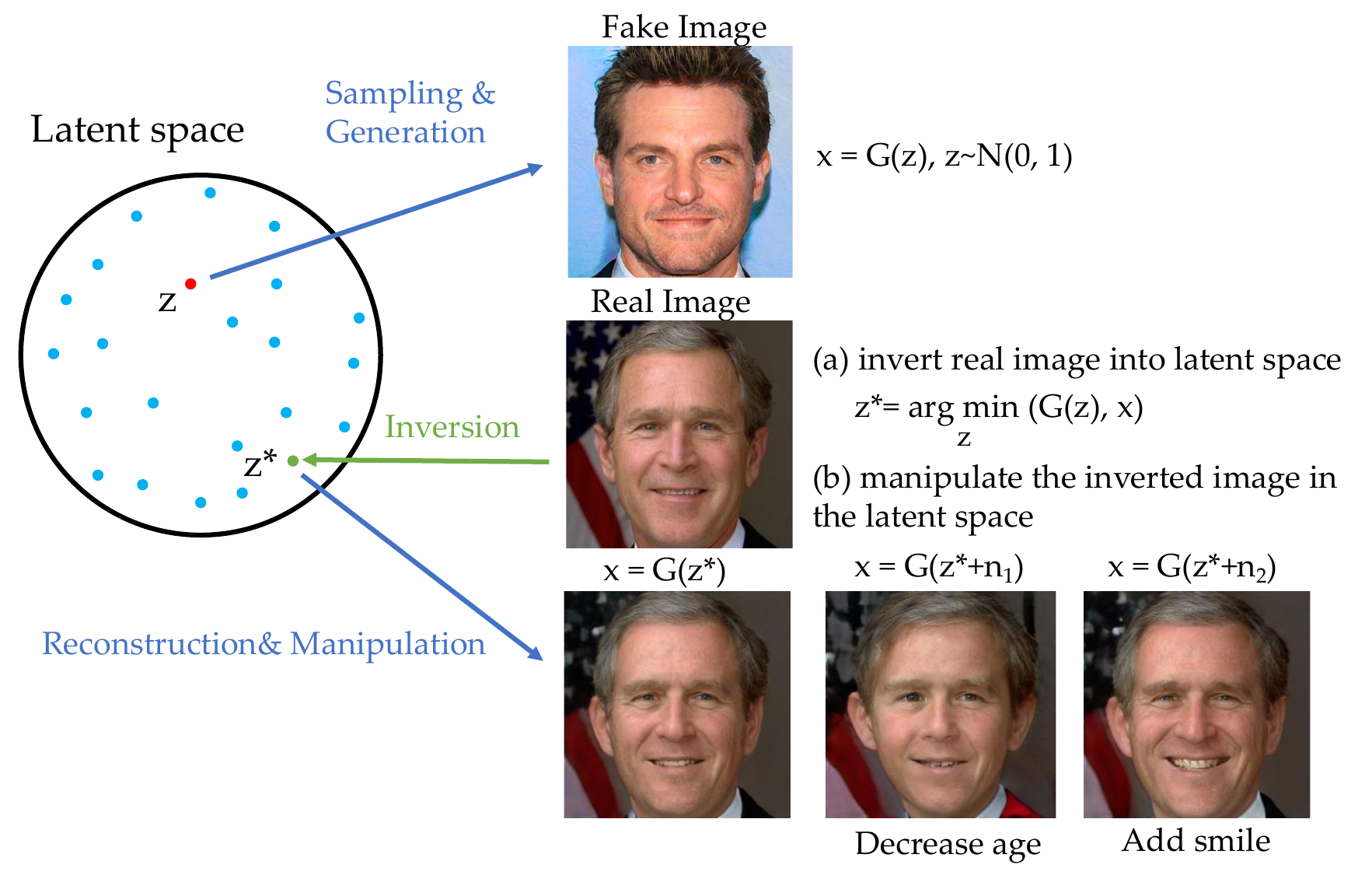}
\end{center}
\caption{{Illustration of GAN inversion.} Different from the conventional sampling and generation process using trained generator $G$, GAN inversion maps a given real image $x$ to the latent space and obtains the latent code $\mathbf{z^{*}}$. The reconstructed image $x^{*}$ is then obtained by $x^*=G(\mathbf{z^{*}})$. By varying the latent code $\mathbf{z^{*}}$ in different interpretable directions \eg, $\mathbf{z^{*}}+\mathbf{n_1}$ and $\mathbf{z^{*}}+\mathbf{n_2}$ where $\mathbf{n_1}$ and $\mathbf{n_2}$ model the age and smile in the latent space respectively, we can edit the corresponding attribute of the real image. The reconstructed results are from \cite{zhu2020indomain}.
}
\label{fig:overview}
\end{figure}
}

\newcommand{\figtype}{
\begin{figure}[tbp]
\begin{center}
\includegraphics[width=0.95\linewidth]{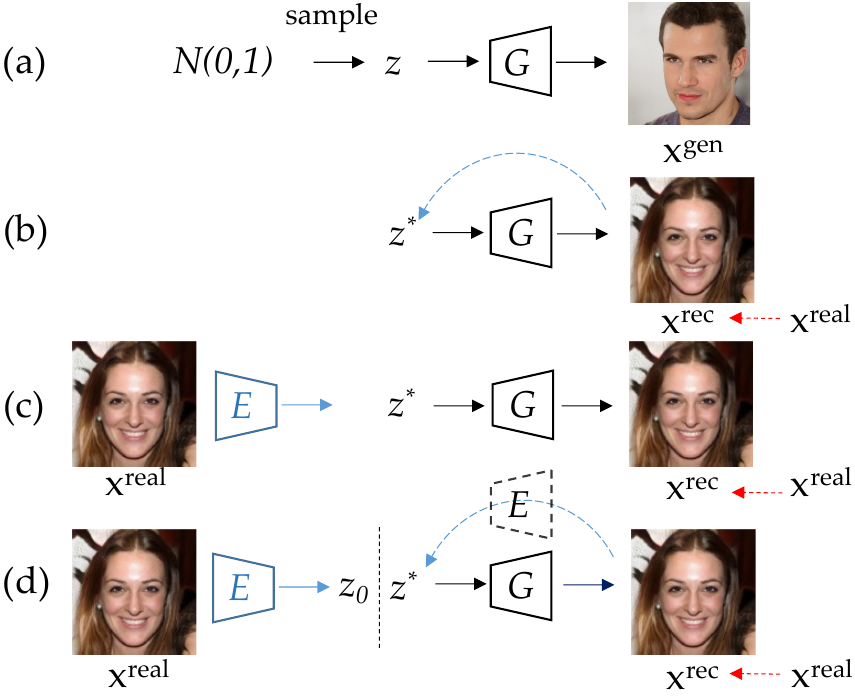}
\end{center}
\caption{Illustration of GAN Inversion Methods.
(a) Given a well-trained GAN model $G$, photo-realistic images $\x^\text{gen}$ can be generated from randomly sampled latent vectors $\z$. GAN inversion aims to obtain the latent code $\z^*$ for a given image $\x^\text{real}$.
A {\bf learning-based} inversion method aims to learn an encoder network to map an image into the latent space such that the reconstructed image based on the latent code look as similar to the original one as possible. 
An {\bf optimization-based} inversion approach directly solves the objective function through back-propagation to find a latent code that minimizes pixel-wise reconstruction loss. 
A {\bf hybrid} approach first uses an encoder to generate initial latent code and then refines it with an optimization algorithm.
Depicted by the dotted $E$, the well-trained encoder is included in \cite{zhu2020indomain} as a regularizer for optimization.
Blue blocks represent trainable or iterative modules, and red dashed arrows indicate the supervisions.}
\label{fig:inversion_types}
\end{figure}
}

\newcommand{\figmodel}{
\begin{figure}[tbp]
\begin{center}
\includegraphics[width=0.95\linewidth]{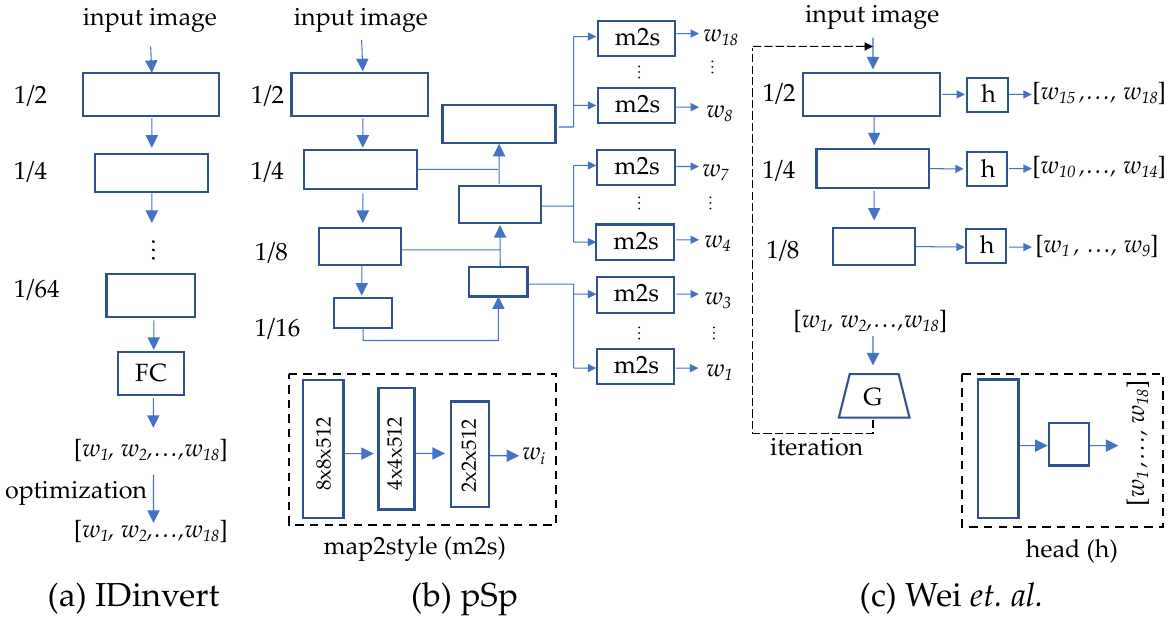}
\end{center}
\caption{
{Encoder structure of three learning-based methods: (a) IDinvert~\cite{zhu2020indomain}, (b) pSp~\cite{richardson2020encoding}, and (c) Wei~\etal~\cite{wei2021simpleinversion}.}
}
\label{fig:modelcompare}
\end{figure}
}

\newcommand{\fignoninference}{
\begin{figure}[tbp]
\begin{center}
\includegraphics[width=0.6\linewidth]{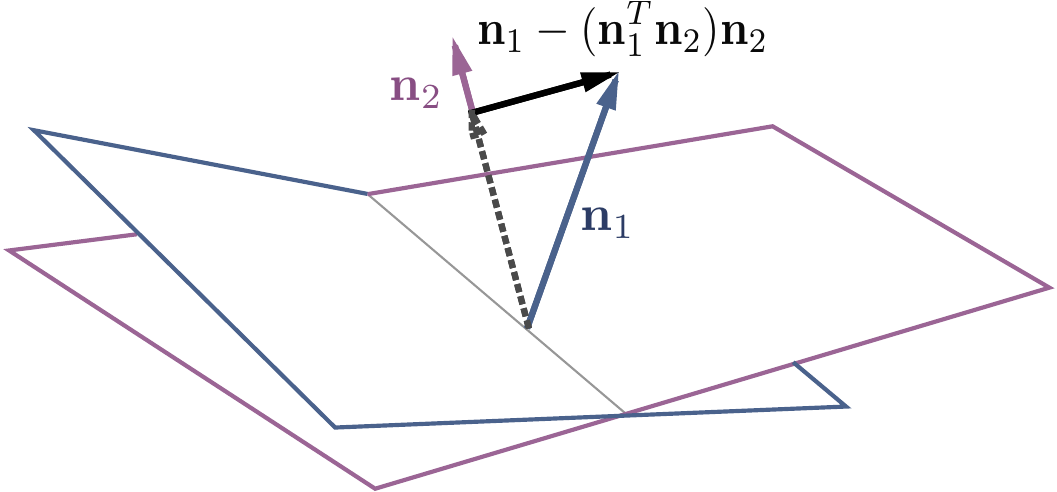}
\end{center}
\caption{Illustration of discovering disentangled directions for multiple attributes.
The projection of $\mathbf{n}_{1}$ onto $\mathbf{n}_{2}$ is subtracted from $\mathbf{n}_{1},$ resulting in a new direction $\mathbf{n}_{1}-(\mathbf{n}_{1}^{\top} \mathbf{n}_{2}) \mathbf{n}_{2}$. This figure is from~\cite{shen2020interpreting}.}
\label{fig:projection}
\end{figure}
}

\newcommand{\figspace}{
\begin{figure}[tbp]
\begin{center}
\includegraphics[width=0.95\linewidth]{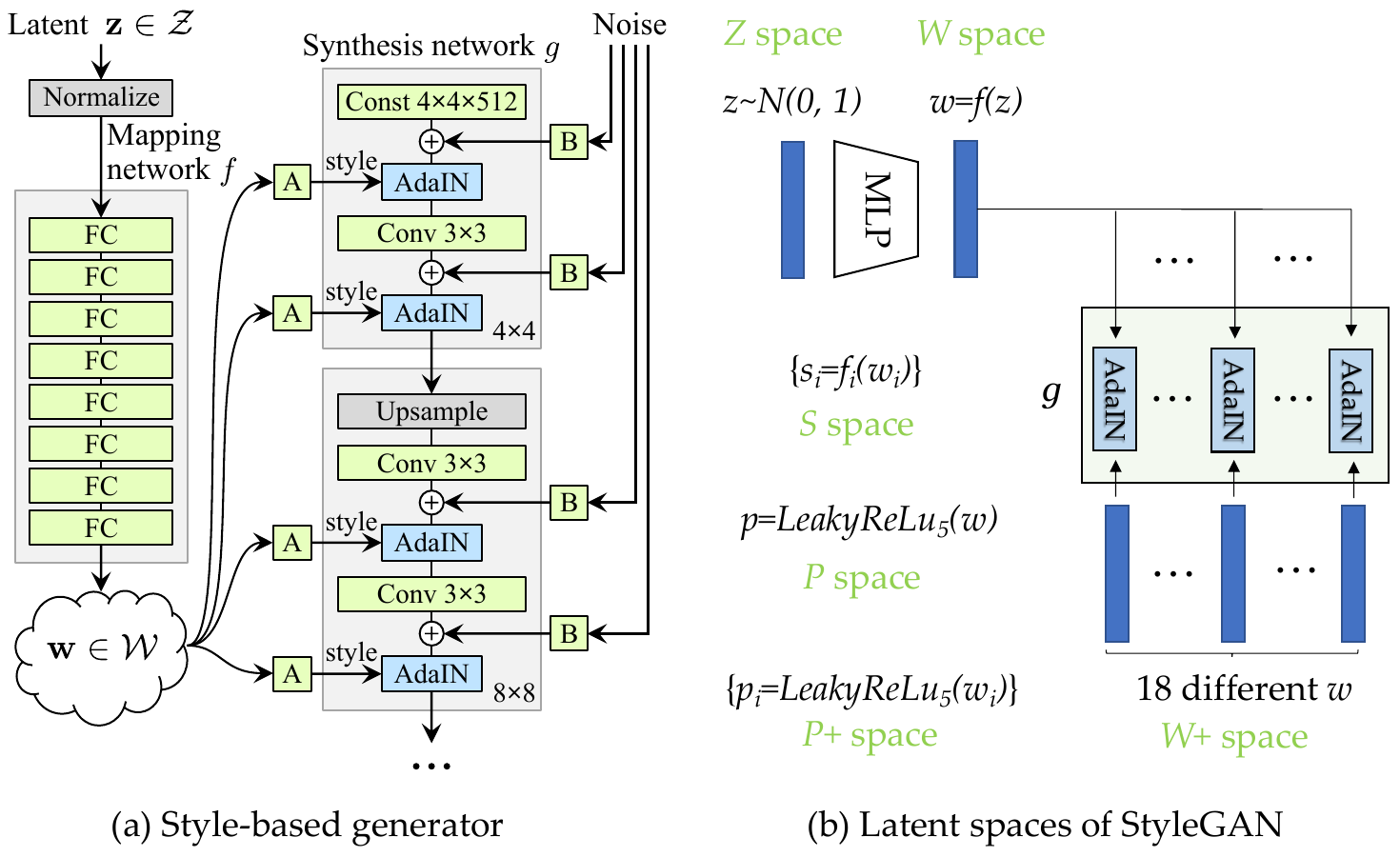}
\end{center}
\caption{(a) Architecture of the style-based generator. (b) The latent spaces from which the inversion methods are constructed.
The synthesis network \textit{g} and AdaIN in (b) are the same as in (a).
}
\label{fig:space}
\end{figure}
}

\newcommand{\tabfeature}{
\begin{table*}[t]
\renewcommand{\arraystretch}{1.3}
\caption{Properties of GAN inversion methods. \textit{Type} includes Learning-based (L.), Optimization-based (O.), and Hybrid (H.) GAN inversion. S.-A., L.-W., and S.-R denote Semantic Awareness, Layerwise, and Supported Resolution, respectively. \textit{GAN model} and \textit{Dataset} indicate which GAN models are trained on which dataset that a method is inverting, which can be found in Section~\ref{sec:model_data}.}
\label{tab:taxonomy}
\rowcolors{2}{gray!0}{gray!5}
\centering
{
\setlength{\fboxrule}{1pt}
\resizebox{\textwidth}{!}{%
\begin{tabular}{|l|c|c|c|c|c|c|c|c|l|}
\hline
Method  & Publication & Type &S.-A. &L.-W. & S.-R & Space &GAN Model & Dataset & Keywords \\\hline
Zhu~\etal~\cite{zhu2016generative} &ECCV'16 &H. & & &64 &$\mathcal{Z}$ &\cite{radford2016dcgan} &\cite{yu2014local,zhou2014places,yu2015lsun} &inversion for GANs\\\hline
Creswell~\etal~\cite{creswell2016inverting,creswell2018inverting} &NeurIPS'16 &O. & & &128 &$\mathcal{Z}$ &\cite{radford2016dcgan,gulrajani2017improved} &\cite{yu2014local,liu2015faceattributes} &first using the term \textit{inversion}
\\\hline
Perarnau~\etal~\cite{perarnau2016invertible} &NeurIPS'16 &L. & & &64 &$\mathcal{Z}$ &\cite{radford2016dcgan} &\cite{lecun1998mnist,liu2015faceattributes} &inversion for conditional GAN\\\hline
GANPaint~\cite{bau2019ganpaint} &TOG'19 &H. &\yes &\yes &256 &$\mathcal{Z}$ &\cite{karras2017progressive} &\cite{yu2015lsun} &learn an image-specific generator\\\hline
GANSeeing~\cite{bau2019seeing} &ICCV'19 &H. &\yes &\yes &256 &$\mathcal{Z}$, $\mathcal{W}$  &\cite{gulrajani2017improved,karras2017progressive,karras2019style}  &\cite{yu2015lsun} &visualization of mode collapse\\\hline
Image2StyleGAN~\cite{abdal2019image2stylegan} &ICCV'19 &O. &\yes &  &1024 &$\mathcal{W}$ &\cite{karras2019style} &\cite{karras2019style} &first inversion for StyleGAN\\\hline
Image2StyleGAN++~\cite{abdal2020image2stylegan2} &CVPR'20 &O. &\yes &\yes &1024 &$\mathcal{W^+}$ &\cite{karras2017progressive,karras2019style} &\cite{karras2017progressive,karras2019style} &\\\hline
mGANPrior~\cite{gu2020image} &CVPR'20 &O. &\yes &\yes &256 &$\mathcal{Z}$ &\cite{karras2017progressive,karras2019style} &\cite{karras2017progressive,karras2019style,yu2015lsun} &multi-code GAN prior\\\hline
Editing in Style~\cite{collins2020uncovering} &CVPR'20 &O. &\yes &  &1024 &$\mathcal{W}$ &\cite{karras2017progressive,karras2019style,karras2020analyzing} &\cite{karras2019style,yu2015lsun} &\\\hline
YLG~\cite{daras2020your} &CVPR'20 &O. & & &128 &$\mathcal{Z}$ &\cite{zhang2019self} &\cite{russakovsky2015imagenet} &attention\\\hline
Huh~\etal~\cite{huh2020transforming} &ECCV'20 &O. &\yes & &1024 &$\mathcal{Z}$ &\cite{brock2018large,karras2020analyzing} &\cite{russakovsky2015imagenet,yu2015lsun,karras2019style} &class-conditional\\\hline
IDInvert~\cite{zhu2020indomain} &ECCV'20 &H. &\yes &\yes &256 &$\mathcal{W^+}$ &\cite{karras2019style} &\cite{karras2019style,yu2015lsun} & in-domain\\\hline
SG-Distillation~\cite{viazovetskyi2020distillation} &ECCV'20 &O. &\yes &\yes &1024 &$\mathcal{W^+}$ &\cite{karras2020analyzing} &\cite{karras2019style} &\\\hline
MimicGAN~\cite{anirudh2020mimicgan} &IJCV'20 &O. & & &64 &$\mathcal{Z}$ &\cite{liu2015faceattributes} &\cite{radford2016dcgan} &for corrupted images\\\hline
Chai~\etal~\cite{chai2021using} &ICLR'21 &L. &\yes &\yes &1024 &$\mathcal{Z}$, $\mathcal{W^+}$ &\cite{karras2017progressive,karras2020analyzing}  &\cite{karras2017progressive,karras2019style,yu2015lsun} &data augmentation \\\hline
pSp~\cite{richardson2020encoding} &CVPR'21 &L. &\yes &\yes &1024 &$\mathcal{W^+}$ &\cite{karras2020analyzing} &\cite{karras2017progressive} &map2style module\\\hline
StyleSpace~\cite{wu2020stylespace} &CVPR'21 &O. &\yes &\yes &1024 &$\mathcal{S}$ &\cite{karras2020analyzing} &\cite{karras2019style,yu2015lsun} &$\mathcal{S}$-space\\\hline
GH-Feat~\cite{xu2021ghfeat} &CVPR'21 &L. &\yes &\yes &256 &$\mathcal{S}$ &\cite{karras2019style} &\cite{lecun1998mnist,yu2015lsun,karras2019style} &generative hierarchical feature\\\hline
GANEnsembling~\cite{chai2021ensembling} &CVPR'21 &H. &\yes &\yes &1024 &$\mathcal{W^+}$ &\cite{karras2020analyzing} &\cite{karras2019style,yu2015lsun}  & \\\hline
e4e~\cite{tov2021designing} &TOG'21 &L. &\yes &\yes &1024 &$\mathcal{W^+}$ &\cite{karras2020analyzing} &\cite{karras2017progressive,karras2019style,yu2015lsun} &encoder for editing\\\hline
Xu~\etal~\cite{xu2021continuity} &ICCV'21 &O. &\yes &\yes &1024 &$\mathcal{W^+}$ &\cite{karras2019style} &\cite{livingstone2018ryerson,karras2019style} & for consecutive images\\\hline
ReStyle~\cite{alaluf2021restyle} &ICCV'21 &L. &\yes &\yes &1024 &$\mathcal{W^+}$ &\cite{karras2020analyzing} &\cite{karras2017progressive,karras2019style,choi2020starganv2,yu2015lsun}  &iterative refinement\\\hline
BDInvert~\cite{kang2021gan} &ICCV'21 &O. &\yes &\yes &1024 &$\mathcal{F/W^+}$ &\cite{karras2017progressive}, \cite{karras2020analyzing} &\cite{karras2017progressive,karras2019style,yu2015lsun} & out-of-range, $\mathcal{F/W^+}$-space  \\\hline
Zhu~\etal~\cite{zhu2020improved} &arxiv'21 &O. &\yes &\yes &1024 &$\mathcal{P}$ &\cite{karras2019style,karras2020analyzing} &\cite{karras2019style} & $\mathcal{P}$ and $\mathcal{P^+}$space \\\hline
Wei~\etal~\cite{wei2021simpleinversion} &arxiv'21 &L. &\yes &\yes &1024 &$\mathcal{W^+}$ &\cite{karras2020analyzing} &\cite{karras2017progressive,karras2019style} &efficient encoder architecture \\\hline
PTI~\cite{roich2021pivotal} &arxiv'21 &H. &\yes & &1024 &$\mathcal{W}$ &\cite{karras2020analyzing} &\cite{karras2017progressive,karras2019style} & tune $G$ around a pivot latent code \\\hline
HyperStyle~\cite{alaluf2021hyperstyle} &CVPR'22 &H. &\yes & &1024 &$\mathcal{W}$ &\cite{karras2020analyzing} &\cite{karras2017progressive,karras2019style,krause20133d} & learn to optimize the generator \\\hline
HFGI~\cite{wang2021high} &CVPR'22 &L. &\yes &\yes &1024 &$\mathcal{W^+}$ &\cite{karras2019style,karras2020analyzing} &\cite{karras2017progressive,karras2019style,krause20133d} &  \\\hline
HyperInverter~\cite{dinh2021hyperinverter} &CVPR'22 &L. &\yes & &1024 &$\mathcal{W}$ &\cite{karras2020analyzing} &\cite{karras2017progressive,karras2019style,yu2015lsun} &  two-phase inversion \\\hline
\end{tabular}
} %
} %
\end{table*}
}